\documentclass[accepted]{uai2026} 
                        
\usepackage[american]{babel}

\usepackage{natbib} 
\usepackage{mathtools} 
\usepackage{booktabs} 
\usepackage{tikz} 
\usepackage{subcaption}

\usepackage{tikz}
\usetikzlibrary{
  positioning,
  bayesnet,
  fit,
  shapes.geometric,
  shapes,
  arrows.meta,
  calc
}
\tikzset{
  rounded box/.style={rectangle, rounded corners, draw, align=center, minimum height=1cm, minimum width=1.5cm},
  startstop/.style={rectangle, rounded corners, minimum width=3cm, minimum height=1cm, text centered, draw=black, fill=red!30},
  process/.style={rectangle, minimum width=3cm, minimum height=1cm, text centered, draw=black, fill=blue!30},
  decision/.style={diamond, minimum width=3cm, minimum height=1cm, text centered, draw=black, fill=green!30},
  arrow/.style={->,>=stealth},
  bidirected/.style={<->,>=stealth},
  dottededge/.style={dotted, thick}
}

\title{How PC-based Methods Err: Towards Better Reporting of Assumption Violations and Small Sample Errors}

\author[1, *]{\href{sofia.faltenbacher@uni-potsdam.de}{Sofia Faltenbacher}{}}
\author[2, *]{\href{jonas.wahl@dfki.de}{Jonas Wahl}{}}
\author[1]{Rebecca Herman}
\author[1]{Jakob Runge}
\affil[1]{University of Potsdam
}
\affil[2]{German Research Center for Artificial Intelligence (DFKI)
}
\affil[*]{Equal contribution
}

\newcommand{\ind}{\rotatebox[origin=c]{90}{$\models$}}

\newcommand{\x}{{\mathbf{x}}}

\newcommand{\Sset}{{\mathbf{S}}}

\newcommand{\X}{{\mathbf{X}}}

\newcommand{\G}{{\mathcal{G}}}
\newcommand{\cG}{{\mathcal{G}}}
\newcommand{\cH}{{\mathcal{H}}}

\newcommand{\cC}{{\mathcal{C}}}
\newcommand{\cM}{{\mathcal{M}}}
\newcommand{\cP}{{\mathcal{P}}}

\newcommand{\cT}{{\mathcal{T}}}

\newcommand{\conditionedon}{\,|\,}

\usepackage[utf8]{inputenc} 
\usepackage[T1]{fontenc}    
\usepackage{hyperref}       
\usepackage{url}            
\usepackage{booktabs}       
\usepackage{amsfonts}       
\usepackage{nicefrac}       
\usepackage{microtype}      
\usepackage{xcolor}         
\usepackage{centernot}
\usepackage{amsthm}

\usepackage[american]{babel}

\usepackage{mathtools}

\theoremstyle{plain}

\newtheorem{lemma}{Lemma}
\newtheorem{proposition}{Proposition}
\newtheorem{observation}{Observation}
\newtheorem{corollary}{Corollary}[proposition]

\theoremstyle{definition}
\newtheorem{definition}{Definition}
\newtheorem{example}{Example}

\theoremstyle{remark}

\begin{document}
\maketitle

\begin{abstract}
Causal discovery methods based on the PC algorithm are proven to be sound if all structural assumptions are fulfilled and all conditional independence tests are correct. This idealized setting is rarely given in real data. In this work, we first analyze how local errors can propagate throughout the output graph of a PC-based method, highlighting how consequential seemingly innocuous errors can become. Next, we introduce coherency scores to find assumption violations and small sample errors in the absence of a ground truth. These scores do not require statistical tests beyond those already executed by the causal discovery algorithm. Errors detected by our approach extend the set of errors that can be detected by comparable existing methods. We place our computationally cheap global error detection and quantification scores as a bridge between computationally expensive global answer-set-programming-based methods and less expensive local error detection methods. The scores are analyzed on simulated and real-world datasets.
\end{abstract}

\section{Introduction}\label{sec:intro}
Causal discovery, also known as causal structure learning, is an active research field with new methods being published at almost every major machine learning conference; see \cite{zanga2022survey,assaad2022survey,brouillard2024landscape} for recent surveys on methods and applications. The field aims to provide tools for inferring qualitative cause-and-effect relationships from observational and experimental data, typically in the form of a graphical model, while elucidating the assumptions underlying such inference. One of the seminal causal discovery algorithms that inspired many others is the PC algorithm \citep{spirtes1991algorithm}.

The correctness of the PC algorithm and its descendants, such as FCI \citep{spirtes2001anytime} and PCMCI \citep{runge_discovering_2020}, was proven for an idealized setting in the absence of finite sample errors. In this setting, it is shown that the output is maximally informative as long as the data generation process satisfies the assumptions of the algorithm \citep{Spirtes2000}. Theoretical results on finite sample performance are much harder to prove and therefore rare; see \cite{kalisch2007estimating}. Assessments of finite sample performance employ any of a variety of performance metrics \citep{tsamardinos_max-min_2006,peters2015structural, henckel2024adjustment,wahl2024separationbased}, but these metrics usually require knowledge of the ground truth causal graph, limiting these evaluations to simulated data. In addition, it is often unclear how well real-world data conforms to the assumptions of causal discovery, leaving users wondering how much they can trust an algorithm's results~\citep{krich2020estimating,miersch2025evaluating}. 

As a consequence of statistical errors and violations of assumptions, PC-based methods may produce an output graph with conspicuous inconsistencies. An example of such an inconsistency is an orientation conflict, a consequence of the algorithm's attempt to orient the same edge in different directions at different points in the procedure. These conflicts show that there is no output graph of the required type that matches the test results of all conditional independence tests executed by the algorithm; otherwise the method would have found it. The consequences of such inconsistencies are often left unaddressed but must be taken seriously as guarantees for downstream tasks like causal effect estimation no longer hold if the provided causal graph is not correct.

Related problems have recently been pointed out by researchers in the field. \cite{jorgensen2025causal} discuss real-world examples where typically observed variables cannot be adequately modeled by an SCM. \cite{miersch2025evaluating} have evaluated the robustness of PC-based time series causal discovery for flood drivers and have found that some key hydrological mechanisms are missed even for very large sample sizes. \cite{gamella2025causal} have pointed out the poor performance of several causal discovery algorithms on physical testbeds. We provide a more detailed discussion in the Appendix \ref{app:recent_critisism}. These works underline the need for rigorous evaluation of causal discovery methods on real-world data, under assumption violations, and in the presence of statistical errors.

\begin{figure}[!htb]
  \centering
  \includegraphics[width=\linewidth]{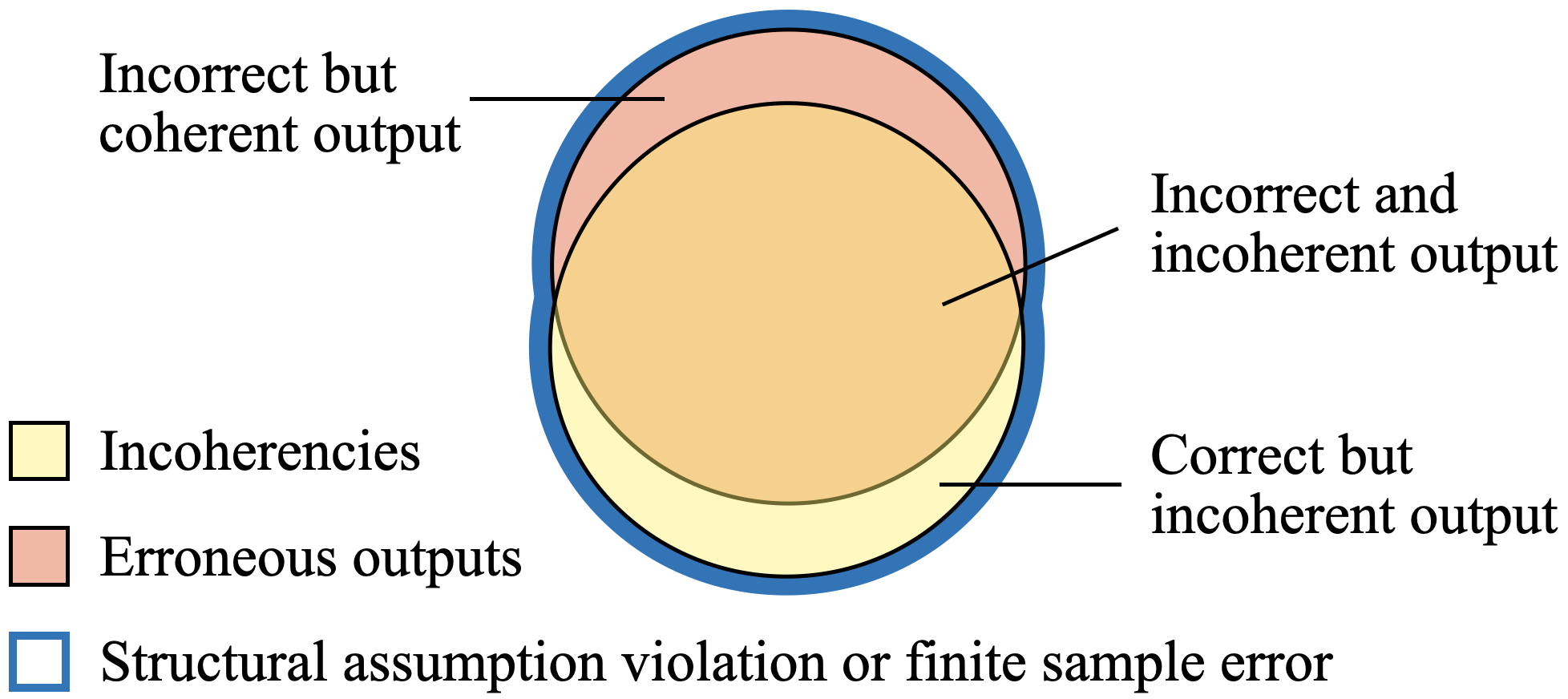}
  \caption{Resolving orientation conflicts and ambiguities always leads to incoherencies, see Corollary \ref{cor:resolving}. Outputs without conflicts and ambiguities can still be incoherent, see Observation \ref{obs:no_conflicts_but_detectable_errors}. Assumption violations or statistical errors that lead to erroneous outputs (orange intersection of red and yellow, see Example \ref{ex:unfaithfulness_detectable}, are a real subset of incoherencies that can lead to correct or incorrect graphs (yellow) and erroneous outputs that can be coherent or incoherent (red). See Example \ref{ex:correct_but_incoherent} for a correct but incoherent output and Example \ref{ex:very_small_effects_undetectable_erroneous} for a coherent but incorrect output.}\label{fig:diagram}
\end{figure}

Currently, there are local checks for ambiguities \citep{ramsey_adjacency-faithfulness_2006, colombo_order-independent_2014} and global but computationally expensive methods based on answer set programming (ASP) \citep{bromberg2009improving, russo2024argumentative, hyttinen2014constraint} to address the inconsistency problem. Further, graph falsification approaches \citep{eulig_toward_2023, ramsey2024choosing, shipley2000new} as well as a method sanity check \citep{faller2024self} consider adjacent topics. We differentiate our method from related work and point out synergies in a detailed comparison in Section \ref{sec:related_work}. In essence, our method may be seen as a bridge between existing local refinements and global but computationally expensive methods. Specifically, our main contributions are as follows:
\begin{itemize}
  \item We introduce a new, computationally cheap coherency score for PC-based methods. In essence, our score measures how well the conditional independence test results obtained during a method's execution match the separations and connections implied by its output. Crucially, the score can be computed without access to a ground truth graph.
  \item We characterize the subset of erroneous output graphs of a PC-based method that are detectable without access to the ground truth and without running additional tests, see Figure \ref{fig:diagram}.  
  \item We show how computing our score on subsets of data helps to distinguish between assumption violations and statistical errors. We demonstrate error analysis with our scores on simulated and real-world data.
  \item We provide extensive experiments showcasing that our score serves as a heuristic proxy for the Structural Hamming Distance (SHD) to the unknown ground truth.
\end{itemize}

\section{Preliminaries and notation}  \label{sec:preliminaries}

In this work, $\X=(X^1, \dots, X^d)$ with $d \in \mathbb{N}$ denotes a set of nodes that represent random variables in a structural causal model (SCM) as defined by \cite{pearl_causality_2009}, and $\cG$ denotes a causal graph over $\X$. The causal graph implied by the underlying true SCM is denoted by $\cG_{true}$. Causal graphs come with a notion of graphical \emph{separation} such as $d$-separation for directed acyclic graphs (DAGs, e.g. \cite{pearl_causality_2009}), $m$-separation for maximal ancestral graphs (MAGs, e.g. \citep{zhang_causal_2008}), and $\sigma$-separation for cyclic graphs (e.g. \citep{bongers_foundations_2021}). Unless indicated otherwise, statements in this work do not depend on the type of graph as long as one consistently uses the corresponding criterion for separation and its counterpart: \emph{connection}. We represent separation and connection statements with tuples consisting of two singular nodes and a (possibly empty) disjoint subset of nodes. The set of all such statements is denoted by
\begin{equation*}
    \cC_{\cG}=\left\{(X,Y,\Sset) \, \big| \, X,Y \in \X, \ X \neq Y, \Sset \subseteq \X \backslash\{X,Y \} \right\},
\end{equation*}
and given a causal graph $\cG$, the relevant notion of separation specifies for each tuple $(X,Y,\Sset)$ whether it is a \emph{separation statement} $X\bowtie_{\cG} Y\conditionedon \Sset$ ($X$ and $Y$ are separated given $\Sset$) or a \emph{connection statement} $X\centernot{\bowtie}_{\cG}Y \conditionedon \Sset$ ($X$ and $Y$ are connected given $\Sset$). We define the separation indicator function $\iota_{\cG}:\cC_{\cG} \to \{0,1 \}$ as $\iota_{\cG}(X,Y,\Sset)=1$ if $X \bowtie_{\cG} Y \conditionedon \Sset$ and as $\iota_{\cG}(X,Y,\Sset)=0$ otherwise.
The set of separations and connections are respectively denoted by
\begin{align*}
    \cC^{sep}(\mathcal{G})&=\left\{(X,Y,\Sset) \in \cC_{\cG} \, \big| \, \iota_{\cG}(X,Y,\Sset)=1 \right\}, \\
    \cC^{con}(\mathcal{G})&=\left\{(X,Y,\Sset) \in \cC_{\cG} \, \big| \, \iota_{\cG}(X,Y,\Sset)=0 \right\}.
\end{align*}

In a causal graph, an \emph{unshielded triple} is a triple of nodes $(X,Y,Z)$ such that the outer nodes are adjacent to the middle node but not to each other. An unshielded triple is a \emph{collider} if the edges $(X,Y)$ and $(Y,Z)$ point towards $Y$ (e.g. $X \to Y \leftarrow Z$), otherwise the triple is called a \emph{non-collider}. 

Two causal graphs $\cG, \cH$ in the same graph class are \emph{Markov equivalent} if they share the same separations, i.e. $\cC^{sep}(\cG) = \cC^{sep}(\cH)$ or, equivalently, the same connections. We denote their Markov equivalence class (MEC) by $[\cG]=\{\cH\,\conditionedon\, \cH\sim_{\mathcal{M}} \cG\}.$
MECs of DAGs can be represented by Complete Partially Directed Acyclic Graphs (CPDAGs), those of MAGs by Partial Ancestral Graphs (PAGs). 


A distribution $\mathcal{P}_{\X}$ over the variables $\X$ is called \emph{Markovian} on a causal graph $\cG$ or an equivalence class $[\cG]$ if the separation statements $X \bowtie_{\cG} Y \conditionedon \Sset$ imply the conditional dependencies $X \ind_{\mathcal{P}_{\X}} Y \conditionedon \Sset$ between the corresponding variables. It is called \emph{faithful} on $\cG$ if the conditional independencies $X \ind_{\mathcal{P}_{\X}} Y \conditionedon \Sset$ imply the separation statements for the corresponding nodes in $\cG$.

$\cM$  will be our symbol for a sound constraint-based causal discovery method. For example, $\cM$ could be a stand-in for (variants of the) PC or FCI algorithm (see, e.g. \cite{spirtes_causation_1993}), the PC-stable algorithm (see, e.g. \cite{colombo_order-independent_2014}), or the PCMCI algorithm family (see, e.g. \cite{runge_discovering_2020}). We denote the output graph of a PC-based method $\cM$ by $\cG_{out}$.

Let $( \x_1,\dots,\x_n ) \in \mathbb{R}^{(d \times n)}$ be $n \in \mathbb{N}$ samples drawn from the $d$-dimensional distribution $\mathcal{P}_{\X}$. Any constraint-based causal discovery method uses one or several statistical tests for conditional independence (CI), or dependence \citep{malinsky2024cautious} for which all statements can be formulated analogously. Let $T$ be a CI test and let $\alpha$ be the significance level. For brevity, we assume that the PC-based method uses only one type of CI test on all tested statements, but all results can be generalized in a straightforward way to algorithm variants using different types of tests. 

Common CI tests include Fisher's Z test for conditional independence in the linear Gaussian setting, Kernel-based methods \citep{zhang2011kernel} or regression-based tests including the generalized covariance measure \citep{shahpeters}. 

We define a conditional independence test indicator function $\iota_{(T,\alpha)}:\cC \to \{0,1 \}$ for a CI test $(T,\alpha)$ as $\iota_{(T,\alpha)}(X,Y,\Sset) =1$ if $X \ind_{(T,\alpha)} Y \conditionedon \Sset$ (i.e. the tuple tested independent), and $\iota_{(T,\alpha)}(X,Y,\Sset) =0$ otherwise.
By $\cT(\cM,T,\alpha) \subseteq \cC_{\cG}$ we denote the set of all statements that were tested during the execution of the algorithm $\cM$ using test $T$ at significance $\alpha$. We simply write $\cT$ if the referenced method, the CI test, and the threshold are fixed. 
The tuples that tested conditionally independent 
and those for which independence was rejected are denoted by
\begin{align*}
    \cT^{ind} &= \left\{ (X,Y,\Sset) \in \cT(\cM,T,\alpha)  \,\big|\,  \iota_{(T,\alpha)}(X,Y,\Sset)=1\right\}, \\
    \cT^{dep} &= \left\{ (X,Y,\Sset) \in \cT(\cM,T,\alpha) \,\big|\,  \iota_{(T,\alpha)}(X,Y,\Sset)=0\right\}.
\end{align*}

\section{Sources and Detectability of Errors}

In this section, we discuss the sources and manifestations of errors and how they affect the output of a sound PC-based algorithm $\mathcal{M}$. 

\subsection{Error Sources}

Soundness proofs of PC-based methods \citep{spirtes_causation_1993, spirtes2001anytime, runge_discovering_2020, mooij2020constraint} rely on a variety of structural assumptions and error-free CI tests, sometimes called \emph{CI oracles}. If an assumption is violated or a CI test result is incorrect, e.g. due to finite sample randomness, soundness is no longer guaranteed. On finite data, CI testing is considered a difficult problem and requires additional assumptions as no CI test with power against any alternative exists \citep{shahpeters}.


\begin{definition}[Erroneous Tuples]
    We call a tuple $(X,Y,\mathbf{S})$ an \textit{erroneous tuple} if the CI test result for $(X,Y,\mathbf{S})$ of a method $\mathcal{M}$ with CI test $(T,\alpha)$ does not match the separation statement of $(X,Y,\mathbf{S})$ in the ground truth causal graph $\mathcal{G}_{true}$, i.e. if 
    \begin{align*}
        &X\bowtie_{\,\,\cG_{true}} Y \conditionedon \mathbf{S} &\text{but} \qquad  &X \centernot{\ind}_{(T,\alpha)} Y \conditionedon \mathbf{S} \quad \text{ or } \\
        &X \centernot{\bowtie}_{\cG_{true}} \, Y \conditionedon \mathbf{S} &\text{but} \qquad &X \ \ind_{\,\,(T,\alpha)} Y \conditionedon \mathbf{S}.
    \end{align*}
    We call a method $\mathcal{M}$ with CI test $(T,\alpha)$ \textit{CI-deviant} on a dataset $\mathcal{D}$ if there is at least one erroneous tuple in $\cT(\cM,T,\alpha)$.
\end{definition}

On real-world data with unknown ground truth, it is usually impossible to determine whether a specific triple is erroneous or not.

\begin{example}
    For the PC algorithm \citep{spirtes_causation_1993}, any of the following issues can lead to CI-deviance: faithfulness violations; incorrect CI test results due to an unlucky draw of samples, low power or the functional assumptions of the CI test not being fulfilled, e.g., using a linear CI test on variables with nonlinear relationships.
\end{example}

 Assume that we generate data from an SCM with a known ground truth graph $\cG_{true}$.
We speak of a \textit{graph-type mismatch} between $\cG_{true}$ and a method $\cM$ if $\cG_{true}$ is not in the class of graphs considered by the method $\cM$. 

\begin{example} \label{ex:graph_type}
    An example of a graph-type mismatch is the application of the PC algorithm (which assumes data from a ground truth DAG) on a set of causally insufficient observed variables. As a minimal example, take two variables $X$ and $Y$ that are confounded by a hidden variable $Z$. No directed edge can represent that $X$ and $Y$ are confounded.
\end{example}


\subsection{Manifestations of Errors}

Before formally defining an erroneous output of a PC-based method, we recap the concepts of marked orientation conflicts and ambiguities in the output graph of a PC-based method and shed light onto the fact that orientation conflicts introduce global uncertainty in the output graph.

\paragraph{Orientation Conflicts}

An orientation conflict occurs if a method $\mathcal{M}$ tries to assign different orientations to the same edge at different stages of the algorithm. Conflicts can be approached in two alternative ways; either by marking conflicts explicitly in the graph, or by forcing the algorithm to make a decision through a \emph{conflict-resolution strategy}, for instance by ranking tested conditional independencies by some heuristic such as the $p$-value \citep{ramsey2016improving}, or by allowing the algorithm to overwrite previous orientations. Clearly, these strategies have their own downsides. Overwriting makes the outcome depend on the variable order, while using $p$-values as a proxy for test reliability is at best a heuristic, as $p$-values are uniformly distributed under the null hypothesis \citep{ramsey2024choosing}. In the presence of conflicts, even seemingly orientable edges may no longer be trustworthy. The appearance of conflicts is valuable information, as conflicts indicate assumption violations or small sample errors, and when employing a conflict-resolution strategy, this information is no longer directly visible in the output graph. The coherency scores that we will introduce in Section \ref{sec:score} offer a middle ground and will be able to capture this information even after applying resolution strategies. At the same time, they can be used to evaluate different conflict-resolution strategies.

We illustrate how orientation conflicts not only reflect local, but also global unreliability of the orientations in the output graph in Example \ref{ex:orientation_conflicts} in the Appendix.

\paragraph{Ambiguities} Some PC-based algorithms, including conservative PC \citep{ramsey_adjacency-faithfulness_2006} and PC with the majority rule \citep{colombo_order-independent_2014}, may additionally mark some unshielded triples as \emph{ambiguous}, indicating that the algorithm refuses to take a decision as to whether a triple is a collider or a non-collider after additional tests. Resolution strategies can also be applied to ambiguities.


\begin{definition}[Erroneous Output]
    We call an output graph $\mathcal{G}_{out}$ of a method $\mathcal{M}$ an \textit{erroneous output} if 
    \begin{enumerate}
        \item it is not Markov-equivalent to the ground truth graph $\mathcal{G}_{true}$ or
        \item there are marked conflicts or ambiguities in $\mathcal{G}_{out}$.
    \end{enumerate}
\end{definition}

\begin{observation} \label{obs:only_ci_deviance}
A method $\mathcal{M}$ can be CI-deviant, but the output graph may still be Markov-equivalent to the unknown ground truth, i.e., \textit{not} produce an erroneous output, see Examples \ref{ambiguity_resolution_incoherent} and \ref{ex:correct_but_incoherent} in the Appendix.\end{observation}

\begin{observation} \label{obs:graph_mismatch_but_correct_output}
A graph-type mismatch does \textit{not} necessarily lead to an erroneous output. For example, the output CPDAG of PC in Example \ref{ex:graph_type} is $X-Y$ which is Markov-equivalent to the ground truth $X\leftrightarrow Y$, see also Example \ref{ex:causal_insufficiency_undetectable} in the Appendix.
\end{observation}

Throughout this work, access to the ground truth $\mathcal{G}_{true}$ is \textit{not} provided as we are interested in error detection in real-world scenarios. Therefore, we cannot directly test whether a result is CI-deviant or there was a graph-type mismatch. We now introduce the concept of incoherencies, only looking at the list of CI tests and the output graph of a method.

\begin{definition}[Incoherence]\label{def:coherency}
  Let $\cM$ be a PC-based method using a CI test $(T,\alpha)$ with output $ \cG_{\mathrm{out}}$ without marked orientation conflicts or ambiguities. 
  We call a tuple $(X,Y,\mathbf{S})$ an \textit{incoherent tuple} if the CI test result for $(X,Y,\mathbf{S})$ of a method $\mathcal{M}$ does not match the separation statement of $(X,Y,\mathbf{S})$ in the method's output graph $\mathcal{G}_{out}$, i.e. if 
    \begin{align*}
        &X\bowtie_{\,\,\cG_{out}} Y \conditionedon \mathbf{S}  &\text{but} \qquad  &X \centernot{\ind}_{(T,\alpha)} Y \conditionedon \mathbf{S} \quad \text{ or } \\
        &X\centernot{\bowtie}_{\cG_{out}} \,\,Y \conditionedon \mathbf{S}  &\text{but} \qquad  & X \,\ind_{\,\,(T,\alpha)} Y \conditionedon \mathbf{S}.
    \end{align*}
We call the output of a PC-based method \textit{incoherent} if there is at least one incoherent tuple.
\end{definition}
Testing for incoherence is a straightforward task: The results of the CI tests can be saved while performing the method, and separation queries can be answered for all common graph types that are outputs of PC-based algorithms such as CPDAGs and MAGs, after potentially resolving marked conflicts and ambiguities. For d-separation queries in CPDAGs and m-separation queries in MAGs, the computational complexity is $\mathcal{O}(n+m)$ where $n$ is the number of nodes and $m$ the number of edges, see, e.g. \cite{perkovic2015complete}.

The following lemma justifies the use of incoherence as a falsification tool. Colloquially speaking, in the absence of structural assumption violations and finite sample errors, a sound method does not show incoherence.

\begin{lemma}[Incoherent and CI-deviant results] \label{lem:incoherent_CI_deviant}
    Assume that there is no graph-type mismatch between the data generation mechanism and the sound PC-based method $\cM$. Then the existence of an incoherent tuple implies the existence of an erroneous tuple, i.e. CI-deviance of $\cM$.
\end{lemma}
All proofs can be found in Appendix \ref{app:proofs}.
\begin{observation} \label{obs:no_conflicts_but_detectable_errors}
    If the output of a PC-based algorithm has no conflicts and ambiguities, the result can still be incoherent.
\end{observation}
\begin{example} \label{ex:unfaithfulness_detectable}
    We look at the following ground truth structural causal model:
    \begin{align*}
        X&=\eta_X,\,\,\,
        &Z&=X+\eta_Z, \\
        W&=Z+\eta_W, \,\,\,
        &Y&=2X-2W+\eta_Y;
    \end{align*}
    with independent, exogenous noise variables $\eta_X, \eta_Z, \eta_W, \eta_Y \sim \mathcal{N}(0,1)$. The corresponding true causal graph is $\cG_{true}$ in Figure \ref{fig:faithfulness_detectable} on the left. The effects along the two directed paths from $X$ to $Y$ negate each other, violating faithfulness. Assuming no other sources of errors, the classical PC algorithm finds the conditional independencies $\cT^{ind}=\{(X,Y,\emptyset), (Z,Y,\{X,W\}), (W,X,\{Z\})\}$; all other tuples are tested dependent. The output graph is depicted in Figure \ref{fig:faithfulness_detectable} on the right. No orientation conflicts arise, but $X$ and $Y$ are conditionally independent while being d-connected in $\cG_{out}$, so that $(X,Y,\emptyset)$ is an incoherent tuple. The criterion to distinguish d-separation and d-connection in CPDAGs can, for example, be found in \cite{perkovic2015complete}.
\end{example}
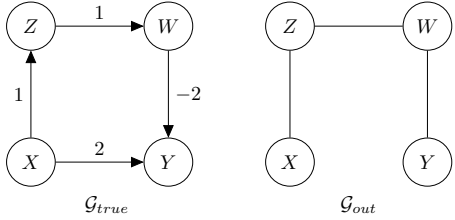
\begin{figure}[h]
    \centering
    \resizebox{.35\textwidth}{!}{ 
        \begin{tabular}{@{}c@{\hspace{1cm}}c@{}} 
            \begin{tikzpicture}[latent/.append style={minimum size=0.85cm}, obs/.append style={minimum size=0.85cm}, det/.append style={minimum size=1.15cm}, wrap/.append style={inner sep=2pt}, on grid]
                \node[latent] (x) at (0,0) {$X$};
                \node[latent] (y) at (2.4,0) {$Y$};
                \node[latent] (z) at (0,2.4) {$Z$};
                \node[latent] (w) at (2.4,2.4) {$W$};
                
                \draw[->] (x) -- (y) node[midway, above] {$2$};
                \draw[->] (x) -- (z) node[midway, left] {$1$};
                \draw[->] (z) -- (w) node[midway, above] {$1$};
                \draw[->] (w) -- (y) node[midway, right] {$-2$};
            \end{tikzpicture}
            &
            \begin{tikzpicture}[latent/.append style={minimum size=0.85cm}, obs/.append style={minimum size=0.85cm}, det/.append style={minimum size=1.15cm}, wrap/.append style={inner sep=2pt}, on grid]
                \node[latent] (x) at (0,0) {$X$};
                \node[latent] (y) at (2.4,0) {$Y$};
                \node[latent] (z) at (0,2.4) {$Z$};
                \node[latent] (w) at (2.4,2.4) {$W$};
                
                \draw[-] (x) -- (z);
                \draw[-] (z) -- (w);
                \draw[-] (w) -- (y);
            \end{tikzpicture}
            \\
            $ \cG_{true} $ & $ \cG_{out} $
        \end{tabular}
    }
    \caption{Faithfulness violation leading to an incoherent erroneous output.}
    \label{fig:faithfulness_detectable}
\end{figure}

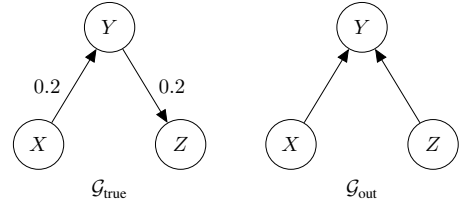
\begin{figure}[h]
    \centering
    \resizebox{.35\textwidth}{!}{ 
        \begin{tabular}{@{}c@{\hspace{1cm}}c@{}} 
            \begin{tikzpicture}[latent/.append style={minimum size=0.85cm}, obs/.append style={minimum size=0.85cm}, det/.append style={minimum size=1.15cm}, wrap/.append style={inner sep=2pt}, on grid]
                \node[latent] (x) at (0,0) {$X$};
                \node[latent] (z) at (2.4,0) {$Z$};
                \node[latent] (y) at (1.2,2) {$Y$};
                
                \draw[->] (x) -- (y) node[midway, left=3pt] {$0.2$};
                \draw[->] (y) -- (z) node[midway, right=3pt] {$0.2$};
            \end{tikzpicture}
            &
            \begin{tikzpicture}[latent/.append style={minimum size=0.85cm}, obs/.append style={minimum size=0.85cm}, det/.append style={minimum size=1.15cm}, wrap/.append style={inner sep=2pt}, on grid]
                \node[latent] (x) at (0,0) {$X$};
                \node[latent] (z) at (2.4,0) {$Z$};
                \node[latent] (y) at (1.2,2) {$Y$};
                
                \draw[->] (x) -- (y);
                \draw[->] (z) -- (y);
            \end{tikzpicture}
            \\
            $ \cG_{\text{true}} $ & $ \cG_{\text{out}} $
        \end{tabular}
    }
    \caption{Comparison of the true causal graph and an undetectable erroneous graph structure.}
    \label{fig:very_small_effects_undetectable_erroneous}
\end{figure}

However, checking for incoherencies is not sufficient to find all erroneous outputs as the next example shows.
\begin{observation} \label{obs:no_conflicts_but_errors}
    If the output of a PC-based algorithm has no conflicts, ambiguities or incoherencies, the result can still be erroneous.
\end{observation}
\begin{example} \label{ex:very_small_effects_undetectable_erroneous}
    We look at the following ground truth structural causal model:
    \begin{equation*}
        X=\eta_X,\,\,\,\,\,
        Y=0.2\cdot X+\eta_Y,\,\,\,\,\,
        Z=0.2\cdot Y+\eta_Z,
    \end{equation*}
    with independent, exogenous noise variables $\eta_X, \eta_Y, \eta_Z \sim \mathcal{N}(0,1)$. The corresponding true causal graph is $\cG_{true}$ in Figure \ref{fig:very_small_effects_undetectable_erroneous}. The total causal effect of $X$ on $Z$ is $0.04$. Assuming no other sources of errors, and choosing the classical PC algorithm as the method $\cM$, the nodes $X$ and $Z$ will be tested independent in most simulations even for 10000 samples, see Section \ref{app:additional_experiments} in the Appendix. Note that the tuple $(X,Z,\{Y\})$ is not tested anymore by $\cM$, as the edge between $X$ and $Z$ has already been removed. Therefore $\cT^{ind}=\{(X,Z,\emptyset)\}$. There are no orientation conflicts, and the method has no internal incoherencies. We have no chance to detect that, in fact, there was an incorrect CI test, and that the output is erroneous without further tests or expert knowledge. A PC-based method $\mathcal{M}$ that also tests $(X,Z,\{Y\})$ (and finds independence), such as parallelized versions, majority- or conservative-PC would detect incoherence in this particular case. Examples where additional CI tests do not help to detect incoherence are discussed in Examples \ref{ex:unfaithfulness_undetectable} and \ref{ex:causal_insufficiency_undetectable} in the Appendix.
\end{example}

Observation \ref{obs:no_conflicts_but_errors} highlights that without access to the ground truth, expert knowledge or additional tests, we cannot in general detect errors of PC-based methods. In other words: There can be output graphs that are coherent but still wrong – and there is no chance to detect it with only the list of CI tests and the output graph of a method. 

We now distinguish three manifestations of CI-deviances or graph-type mismatches on the distribution level (D1-D3) and on the graph level (G1-G3), leading to our main results on detectability of errors without access to a ground truth. Again, we assume that we ran a method $\cM$ on the set of variables $\X$, which found the independencies $\cT^{ind}$, the dependencies $\cT^{dep}$ and produced the output graph $\cG_{out}$. The following three cases are mutually exclusive.

\paragraph{Case-D1: There is no distribution $\cP_{\mathbf{X}}$ that can represent the observed conditional independence test results.} More formally, there is no joint distribution $\cP_{\mathbf{X}}$ on $\X$ such that $(X,Y,\Sset) \in \cT^{ind}$ implies $X \ind_{\cP_{\mathbf{X}}} Y \,|\, \textbf{S}$ and $(X,Y,\Sset) \in \cT^{dep}$ implies $X \centernot{\ind}_{\cP_{\mathbf{X}}} Y \,|\, \textbf{S}$. For instance, if the measured test results violate basic logical properties of conditional independence relations such as the \emph{semi-graphoid axioms} \citep{pearl_causality_2009}, then no distribution $\cP_{\mathbf{X}}$ can accommodate them.

\paragraph{Case-D2: A distribution $\cP_{\mathbf{X}}$ can represent the conditional independence test results, but none is Markovian and faithful to $\cG_{out}$.} Figure \ref{fig:detectable_causal_insufficiency} shows an example of this situation as the result of a causal sufficiency violation.

\paragraph{Case-D3: A distribution $\cP_{\mathbf{X}}$ can represent the conditional independence test results and is Markovian and faithful to $\cG_{out}$.} In this case, we cannot detect that the output is erroneous given the performed CI test results and the output graph, see Example \ref{ex:very_small_effects_undetectable_erroneous} as well as Examples \ref{ex:unfaithfulness_undetectable} and \ref{ex:causal_insufficiency_undetectable} in the Appendix.

The output of a PC-based method belongs to one of the following three categories.

\paragraph{Case-G1: Orientation conflicts or ambiguities are marked in $\cG_{out}$.} Conflicts or ambiguities are only marked if there was an assumption violation or an incorrect CI test result, as the algorithm is proven to be sound otherwise.

\paragraph{Case-G2: No conflicts or ambiguities are marked in $\cG_{out}$, but the results of the PC-based method are incoherent.} An example of this case was discussed in Example \ref{ex:unfaithfulness_detectable}.

\paragraph{Case-G3: No conflicts or ambiguities are marked in $\cG_{out}$ and the results of the PC-based method are coherent.} This case was illustrated in Example \ref{ex:very_small_effects_undetectable_erroneous}, see also Examples \ref{ex:unfaithfulness_undetectable} and \ref{ex:causal_insufficiency_undetectable} in the Appendix.

The following result characterizes the subsets of undetectable erroneous outputs. All proofs of the following results can be found in Section \ref{app:proofs} in the Appendix.

\begin{proposition}[Characterization of Undetectable Erroneous Outputs] \label{prop:G3isD3}
    Erroneous outputs of a method $\cM$ have no marked conflicts and ambiguities and are coherent with the CI test results $\cT^{ind} \cup \cT^{dep}$ (Case-G3) if and only if there is a distribution that can represent the conditional independencies $\cT^{ind}$ and is Markovian and faithful to $\G_{out}$ (Case-D3).
\end{proposition}

\begin{corollary}[Resolved Conflicts and Ambiguities Imply Incoherencies] \label{cor:resolving}
    Resolving conflicts or ambiguities always leads to incoherencies.
\end{corollary}

We have therefore shown that given the list of CI tests of a method $\mathcal{M}$ and its output graph, it is not possible to detect more CI-deviations and graph-type mismatches than the set of incoherent results. This is the theoretical limit to error detection in this setting. The results are summarized and visualized in Figure \ref{fig:diagram}.

\section{Computing coherency}  \label{sec:score}

We quantify the overall coherency of a method by introducing with a coherency score. 

\begin{definition}[Coherency Score]
    Let $\cM$ be a PC-based method using a test $T$ with threshold $\alpha$. Let $w: \cT \rightarrow [0,\infty)$ be a non-trivial weight function. The \textit{coherency score} is defined as
\begin{equation*}
    sc(w,\cM) = 1-\frac{\sum_{(X,Y,\mathbf{S}) \in \cT} w(X,Y,\mathbf{S}) \, \text{inc}(X,Y,\mathbf{S})}{\sum_{(X,Y,\textbf{S}) \in \cT} w(X,Y,\mathbf{S}) },
\end{equation*}
where $\text{inc}(X,Y,\mathbf{S})=|\iota_{\cG}(X,Y, \mathbf{S}) - \iota_{(T,\alpha)}(X,Y,\mathbf{S})|$. Choosing $w(X,Y,\mathbf{S})=1$ yields the \textit{total coherency score}. 
\end{definition}

If the total coherency score is less than one, an incoherency is detected. Given the previous results, this allows us to find all CI-deviances or graph-type mismatches detectable without further information, i.e. only given the list of CI test results performed by $\mathcal{M}$ and its output graph $\mathcal{G}_{out}$. To our best knowledge, we are the first to provide a method to do so.

Different weight functions, i.e., scores other than the total coherency score, are discussed in Appendix \ref{app:weight}. Guidelines on when a score may be considered too low on a particular dataset are provided in Appendix \ref{app:further_tests}.

\section{Experiments}
All details on the implementations used as well as runtimes for the experiments can be found in Appendix \ref{app:additional_experiments}.

\paragraph{Falsification and error analysis} The auto-mgp data set \citep{auto_mpg_9} containing technical specs of cars is used to illustrate the PC algorithm as it yields no orientation-conflicts using the default version (Fisher's Z test, threshold 0.05), for example in the documentation of the popular causal inference Python library \textit{dowhy}. We show that there are incoherencies (total coherency score 0.917) that do not show as orientation conflicts. This implies that when using PC on the auto-mpg dataset, there is a CI-deviance or a graph-type mismatch, i.e., no DAG exists that can represent the CI tests performed by the method. All details including the output graph can be found in Appendix \ref{app:additional_experiments}. For results on the cellular signaling dataset from \cite{sachs_causal_2005}, also see Appendix \ref{app:additional_experiments}.

\paragraph{Simulational studies overview} 

We provide the results of applying the total coherency scores for the PC algorithm on randomly generated DAGs by the structural Hamming distance between the output CPDAG and the CPDAG of the ground truth DAG to illustrate that our scores relate to the distance to the ground truth. We generated data sets from 10400 random DAGs with random edge weights, 1000 samples each and varying numbers of nodes and edge densities. The results are summarized in Figure \ref{fig:proxy}. We further discuss limitations, details on the experimental setup and runtime in Appendix \ref{app:additional_experiments}.

\begin{figure}[!htb]
  \centering
  \includegraphics[width=\linewidth]{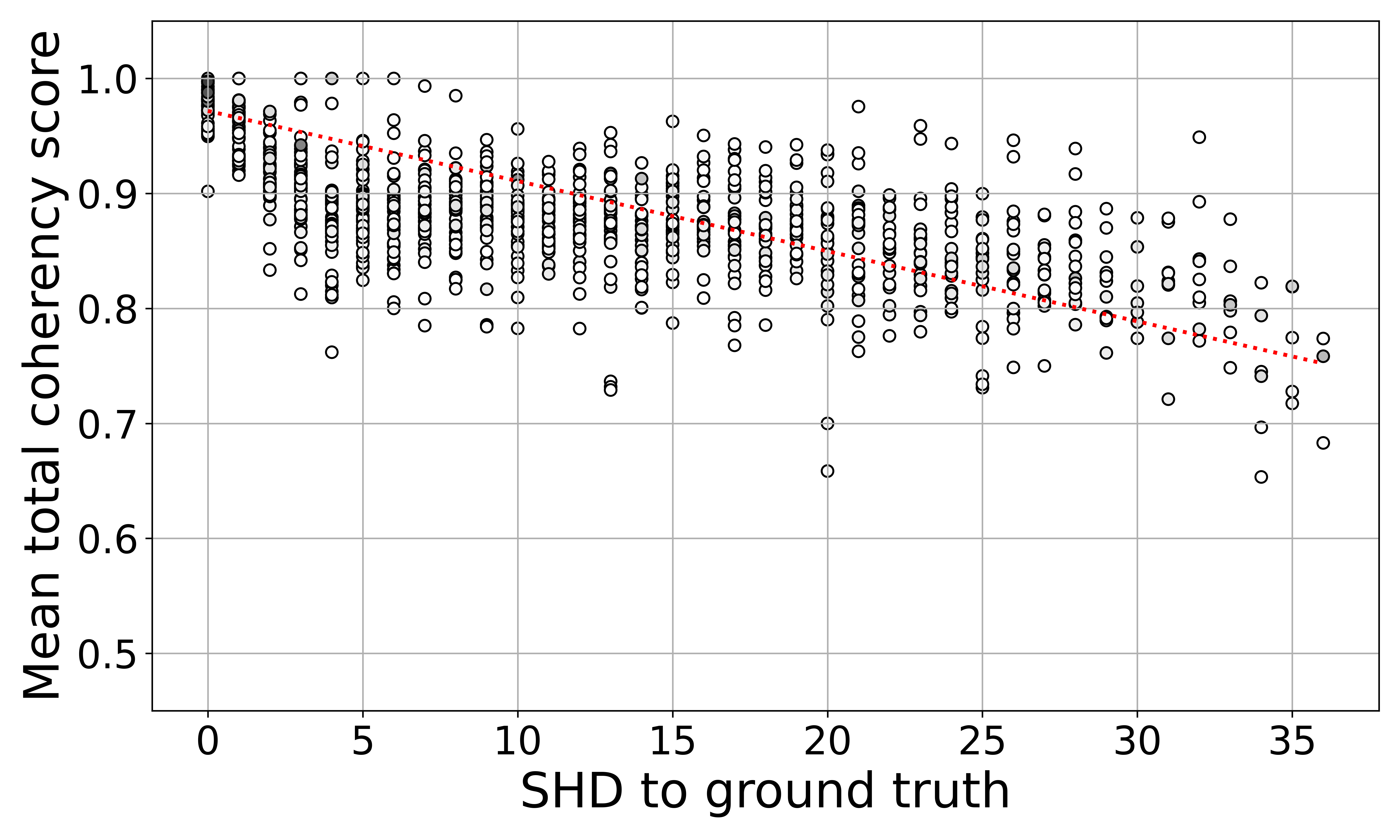}
  \caption{The total coherency scores computed without access to the ground truth serves as a heuristical proxy for the SHD to the ground truth graph.}\label{fig:proxy}
\end{figure}

\paragraph{Assumption violations} If no If all assumptions for the soundness of a method $\mathcal{M}$ are fulfilled, the score reaches one as the sample size increases. For models with detectable assumption violations, i.e., the subset of assumption violations that manifest themselves as incoherencies given oracle CI statements, the score remains at a score smaller than one in theory and simulations. We illustrate the observation in Figure \ref{fig:assumption_analysis}. All details on the models and further simulations can be found in Appendix \ref{app:additional_experiments}. Although it is a difficult problem to distinguish a low from a high score or a plateau from a slow convergence, the observation motivates heuristics to distinguish between assumption violations and statistical errors using our scores on increasing subsets of a given dataset described in the Appendix \ref{app:further_tests}.

\begin{figure}[!htb]
  \centering
  \includegraphics[width=\linewidth]{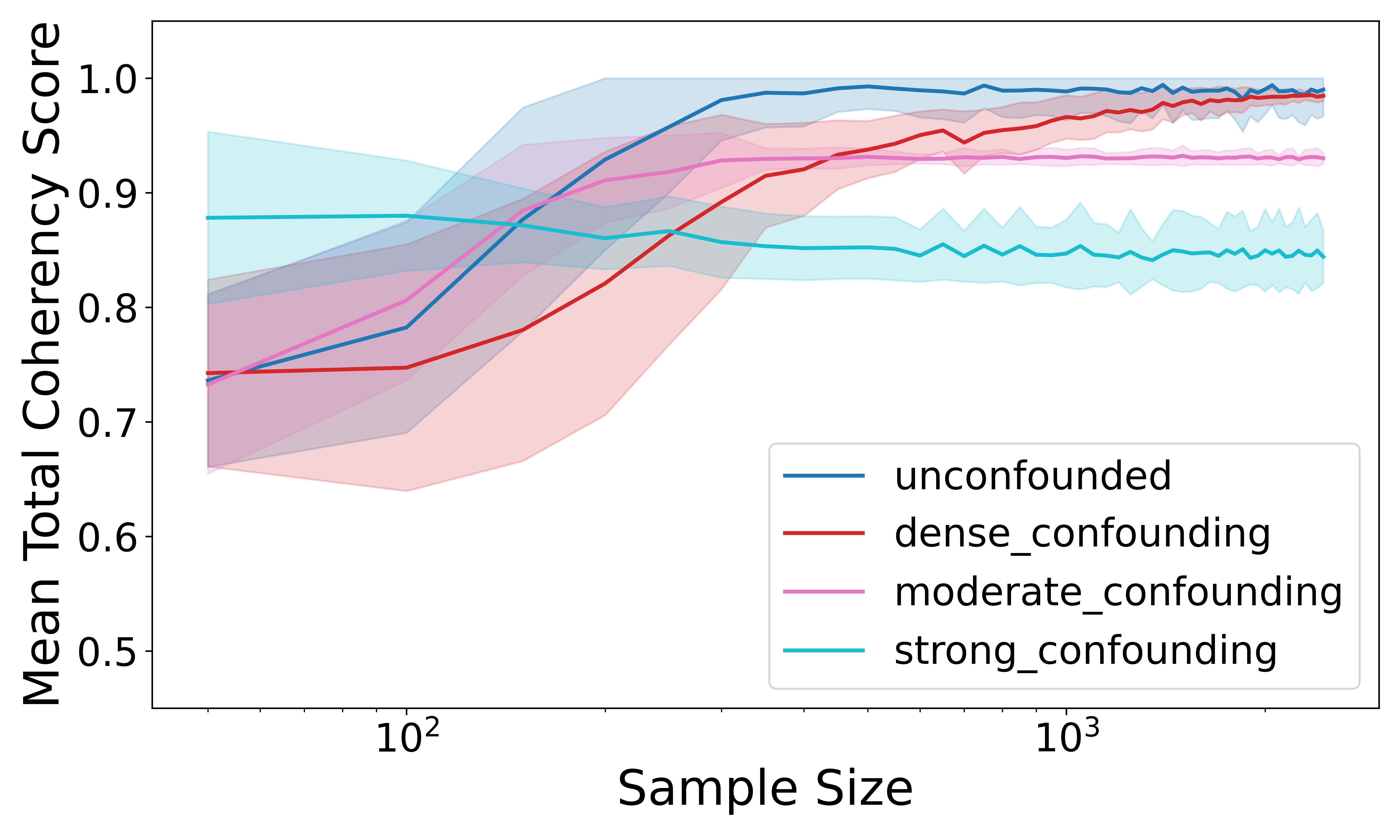}
  \caption{The mean total coherency score for a causal structure without assumption violations (dark blue) vs. for the same structure with different degrees of hidden confounding (red, pink, light blue). The scores reach a plateau at a score smaller across sample sizes for a wide range of assumption violation models.}\label{fig:assumption_analysis}
\end{figure}

\paragraph{Method selection} Consider the following SCM with unobserved variables:
    \begin{align*}
    U_1&=\eta_{U_1}, \quad U_2=\eta_{U_2},\quad &U_3&=\eta_{U_3},\quad U_4=\eta_{U_4} \\
        X&=U_1+U_2+\eta_{X}, \quad &Y&=U_2+U_3+\eta_{Y},\\ Z&=U_3+U_4+\eta_{Z}, \quad &V&=U_4+U_1+\eta_{V}.
    \end{align*}
    where $\eta_{X}, \eta_{y}, \eta_{Z}, \eta_{V}, \eta_{U_1}, \eta_{U_2}, \eta_{U_3}, \eta_{U_4} \sim \mathbb{N}(0,1)$ are independent exogenous noise variables and $U_1,U_2,U_3,U_4$ unobserved. 

While oracle-FCI has a total coherency score of 1, oracle-PC with any possible orientation conflict resolution strategy yields a total coherency score of $\frac{15}{16}$, see Figure \ref{fig:detectable_causal_insufficiency} Appendix \ref{app:additional_experiments}. This allows us to favor oracle-FCI over oracle-PC. Details on the oracle version of the algorithms and the same example with simulated data and CI tests instead of oracle, i.e. true, CI statements can also be found in Appendix \ref{app:additional_experiments}.

\section{Distinction from and Synergies with Related Work} \label{sec:related_work}

The large number of broadly related methods, which differ subtly but significantly in terms of objectives, prerequisites, and practicability, requires a detailed comparison.

\subsection{Refinements of constraint-based algorithms}
In the following, we will discuss the strengths and limits of existing tools to increase robustness and consistency in the context of causal discovery and causal structure learning.

\paragraph{Conservative- and majority-PC} To remedy some of the pitfalls of the PC algorithm in practice, e.g. the order dependence in the presence of orientation conflicts, the variants conservative-PC \citep{ramsey_adjacency-faithfulness_2006} and majority-PC \citep{colombo_order-independent_2014} have been introduced to increase robustness and trust in the result. Both methods essentially add additional CI tests to find ambiguous triples, i.e. unshielded triples that may or may not be oriented as colliders depending on which CI test results one trusts. We have shown that given this new extended list of CI tests compared to the classical PC algorithms, incoherent results are a real superset of results with ambiguous triples; see Obervation \ref{obs:no_conflicts_but_detectable_errors} and Example \ref{ex:unfaithfulness_detectable}.

\paragraph{Argumentative causal discovery} There have been advances in making constraint-based causal discovery algorithms more consistent globally using argumentation frameworks, i.e., logical frameworks to test whether and how severely statements and deductions of statements contradict each other. Argumentation frameworks in causal discovery are used to identify CI statements that should be disregarded to find a set of CI tests that is consistent in itself and/or matches a graph of the type required by the algorithm. Two noteworthy approaches in this area are:
\begin{itemize}
    \item \cite{bromberg2009improving} identify tests that are most contradictory with the knowledge base of CI tests and further CI tests that can be derived from them via the (semi-)graphiod axioms.
    \item \cite{russo2024argumentative} identify tests that are most contradictory with the prerequisites of the PC algorithm, in particular a DAG as ground truth and a match between the d-separations and the list of CI tests.
\end{itemize}
Both methods provide a more elaborate inconsistency resolution than the conflict-resolution strategies implemented in common causal discovery packages such as keep first, keep last or sorting by p-value \citep{ramsey2016improving}. However, there are also several limitations:
\begin{enumerate}
    \item[1.] They both require further assumptions or heuristics. E.g. both methods need an ordering of CI tests by their trustworthiness which both implement by sorting by p-value which also both papers point out as a heuristic without formal guarantees. 
    \item[2.] Both are infeasible for large numbers of variables due to high computational expenses, \cite{russo2024argumentative} explicitly mention 10 variables as a limit.
    \item[3.] We argue that reporting how many CI test results both methods had to disregard in order to find a consistent subset of all performed tests is valuable information which should be reported. This is where we see a synergy when combined with our approach, shedding light on the fact that errors might not only imply that some tests are incorrect due to an unlucky draw of samples. It can also be due to an assumption violation in which case the method is not fitted for the dataset at hand. The rate of tests that were disregarded to obtain a consistent result is an important information to judge how much we trust the output graph.
\end{enumerate}
We also note that for \cite{bromberg2009improving} a consistent set of CI tests does not imply the existence of a matching graph of the type required by the method $\mathcal{M}$, see Case-D2, illustrated by an example in Figure \ref{fig:detectable_causal_insufficiency}. 

\paragraph{Optimization via ASP solvers} \cite{hyttinen2014constraint} approach the inconsistency problem in a puristic way. They take a list of independence and dependence statements as an input, which they choose to be all possible tests on a given dataset. Then, they solve a minimization problem to find the representative graph that minimizes the sum of the weights of the given
conditional independence and dependence constraints that are not implied by the representative graph. They achieve high accuracy and theoretical guarantees. On the downside, this approach explodes in complexity: they report that for seven variables the method takes up to half an hour and for eight variables many instances run for several hours. Like our method, this is a global approach to the inconsistency problem with a different goal: They find the most consistent graph while we report incoherencies as a sanity check of output of a method. Using SGS \citep{Spirtes2000} as the PC-based method $\mathcal{M}$ and their weight function which can be computed categorical or linear Gaussian variables as our weight function, coherency scores can be seen a measure of trust in their results on a single instance. However, our coherency scores can also be used as a sanity check for methods with more scalable runtimes. The score itself only takes less than a second to be computed on an instance for eight variables.

\subsection{Testing graphs against a data set}

There are several tools available to check how well an (expert) given graph or quantitative expert knowledge \citep{grunbaum2023quantitative, eulig_toward_2023, ramsey2024choosing, shipley2000new}. This is different from our setting in which we evaluate the coherence of a method, i.e. whether a method contradicts itself indicating assumption violations or statistical errors to sanity check a method, not a graph. However, some graph falsification methods can be applied to graphs that are the output of a causal discovery algorithm and we further discuss them in Appendix \ref{app:test_graphs}.

\subsection{Method sanity check without ground truth or expert knowledge}
 \cite{faller2024self} check whether the causal graphs learned when a causal discovery algorithm is applied to different subsets of the variables are compatible and introduce different notions of compatibility. Their method is applicable to a wide range of methods including score-based approaches. Their score also serves as a heuristical proxy for the SHD to the unknown ground truth and they also remark that the output graph might be Markov-equivalent to the ground truth despite the output on a subset being incompatible, which, however, still reduces trust in the result. This can be seen as a reflection of the theoretical limits to the evaluation of causal discovery methods without access to the ground truth. 

While our sanity check does not rely on any additional CI tests or causal discovery method runs, they perform the whole causal discovery algorithm with (super-)exponential worst case runtime on several subsets of the variables. We recommend a combined effort. Their computationally more expensive score for score-based methods which ours does not cover, see also see Appendix \ref{app:score_vs_constraint}. Both methods whenever their runtime is not too expensive and/or their particular notions of compatability are of interest. And ours whenever performing PC-based methods as a cheap sanity check with formal guarantees.

\section{Discussion and outlook}

We have shown that testing for incoherencies is an exhaustive error detection approach given only the information collected in one run of the PC-based algorithms and without access to the ground truth. We highlight the potential for error analysis. Similar to many criteria for statistical model selection such as AIC, our scores provide a simple heuristic for a PC-based algorithm's internal coherency. They can be used to choose hyperparameters that score best. We hope that the computationally cheap coherency scores to detect and quantify detectable errors will be reported whenever using PC-based methods on real-world data.

\paragraph{Limitations} While a score smaller than one implies there was a CI-deviance or a graph-type mismatch, the resulting output graph might still be Markov-equivalent to the unknown ground-truth graph. On the other hand, a score of one does not imply that the output graph is correct as some CI-deviances and graph-type mismatches are undetectable. We therefore stress that the score can only serve as a heuristical proxy.

It may be hard to judge when a score is considered too low and the output should be discarded. This can be approached but not resolved by weight function choices and reference values for the output structure. This is not a flaw of our method in particular, but, as we have proven, a theoretical limit for error detection and quantification in this setting.

\bibliography{uai2026-template}

\newpage

\onecolumn

\title{How PC-based Methods Err: Towards Better Reporting of Assumption Violations and Small Sample Errors\\(Supplementary Material)}
\maketitle

\appendix
\section{Proofs} \label{app:proofs}
\textbf{Proof of Lemma \ref{lem:incoherent_CI_deviant}}
By the soundness of PC-based methods, the true causal graph $\mathcal{G}_{true}$ corresponding to the ground truth data generating mechanism is recovered if all assumptions are fulfilled and all CI tests are correct. The existence of an incoherent tuple implies there is no graph $\mathcal{G}$ of the type required by the method such that all CI test results match the separation and connection statements implied by $\mathcal{G}$. Otherwise, it would be the output by the soundness of the method. Therefore, if the graph type required by the method matches the graph type of $\mathcal{G}_{true}$, at least one tuple must be erroneous, i.e. at least one CI test result does not match the separation or connection statement implied by $\mathcal{G}_{true}$.

\textbf{Proof of Proposition \ref{prop:G3isD3}}

\begin{proof}
D1 to D3 and G1 to G3 each partition the set of CI-deviances. We prove that D3 $=$ G3. 

$\subseteq$: D3 is ensured to be a subset of G3 by the soundness of the PC-like algorithm: If given the CI test results, there is a distribution that is Markovian and faithful to $G_{out}$, then the equivalence class of the graph will correctly be found by the PC-like algorithm, in particular there will be no conflicts, ambiguities or incoherencies. 

$\supseteq$: We prove that G3 is a subset of D3 by showing that D1 $\cap$ G3 $=\emptyset$ and D2 $\cap$ G3 $=\emptyset$.

    {D1 $\cap$ G3 $=\emptyset$:} D1 states that no distribution can represent the CI test results $\cT^{ind}\cup\cT^{dep}$. The equivalence class represented by $\cG_{out}$ represents a set of separation statements $\cC_{out}^{sep}$ such that a corresponding set of conditional independence statements can be represented by a family of distributions. As the CI test results cannot be represented by a distribution, this one-to-one correspondence must be violated for at least one tuple $(X,Y,\mathbf{S})$, i.e. 
    \begin{equation*}
        \iota_{(T,\alpha)}(X,Y,\mathbf{S}) \neq \iota_{\cG}(X,Y,\mathbf{S}).
    \end{equation*}

    {D2 $\cap$ G3 $=\emptyset$:} D2 states that the distribution that represents the conditional independencies $\cT^{ind}$ is not Markovian and faithful to any graph in the equivalence class represented by $\cG_{out}$. By definition, that means for at least one tuple $(X,Y,\mathbf{S})$,
    \begin{equation*}
        \iota_{(T,\alpha)}(X,Y,\mathbf{S}) \neq \iota_{\cG}(X,Y,\mathbf{S}).
    \end{equation*}
    Together, D3 $=$ G3.
\end{proof}

\textbf{Proof of Corollary \ref{cor:resolving}}

\begin{proof}
    Proposition \ref{prop:G3isD3} shows that if there are marked conflicts or ambiguities, there is no graph of the type of graph assumed by the method $\mathcal{M}$ that can represent the results of the CI test. Therefore, resolving all conflicts and ambiguities leads to incoherent results, i.e., a mismatch between CI test results and seperation statements.
\end{proof}

\section{Scores for different weight functions} \label{app:weight}

We explicitly formulate the scores for different weight functions used in the paper and introduce additional interesting weight functions.

\begin{itemize}
\item \textbf{(Standard) Independence-Connection-Mismatch.} The weight function
\begin{equation*}w(X,Y,\mathbf{S})=
    \begin{cases}
        0 \text{ if } (X,Y,\Sset) \in \cT^{dep} \\
        1 \text{ else,}
    \end{cases}
\end{equation*}
yields the Independence-Connection-Mismatch:
\begin{equation*}
    r_{ICM}(w,\cM) = 1-\frac{\sum_{(X,Y,\mathbf{S}) \in \cT^{ind}} \left(1-\iota_{\cG}(X,Y,\mathbf{S})\right)}{\left|\cT^{ind} \right|}.
\end{equation*}

\item \textbf{(Standard) Dependence-Separation-Mismatch.} The weight function
\begin{equation*}w(X,Y,\mathbf{S})=
    \begin{cases}
        0 \text{ if } (X,Y,\Sset) \in \cC^{con} \\
        1 \text{ else,}
    \end{cases}
\end{equation*}
yields the Dependence-Separation-Mismatch:
\begin{equation*}
    r_{DSM}(w,\cM) = 1-\frac{\sum_{(X,Y,\mathbf{S}) \in \cC^{sep}} \left(1-\iota_{(T,\alpha)}(X,Y,\mathbf{S})\right)}{\left|\cC^{sep} \right|}.
\end{equation*}

\item \textbf{Conditioning-Set-Size-Adjusted Total Coherency Score.} The weight function
\begin{equation*}
    w(X,Y,\mathbf{S})=\text{exp}(-|\mathbf{S}|)
\end{equation*}
yields the Conditioning-Set-Size-Adjusted Total Coherency Score..

\item \textbf{Conditioning-Set-Size-Adjusted Independence-Connection-Mismatch.} The weight function
\begin{equation*}w(X,Y,\mathbf{S})=
    \begin{cases}
        0 \text{ if } (X,Y,\Sset) \in \cT^{dep} \\
        \text{exp}(-|\mathbf{S}|) \text{ else,}
    \end{cases}
\end{equation*}
yields the Conditioning-Set-Size-Adjusted Independence-Connection-Mismatch.

\item \textbf{Conditioning-Set-Size-Adjusted Dependence-Separation-Mismatch.} The weight function
\begin{equation*}w(X,Y,\mathbf{S})=
    \begin{cases}
        0 \text{ if } (X,Y,\Sset) \in \cC^{con} \\
        \text{exp}(-|\mathbf{S}|) \text{ else,}
    \end{cases}
\end{equation*}
yields the Conditioning-Set-Size-Adjusted Dependence-Separation-Mismatch.
    \item \textbf{Path Length Adjusted Coherency Score.} We can compute a path length adjusted faithfulness coherency score by choosing 
    \begin{equation*}
        w(X,Y,\mathbf{S})=
            \exp(-d(X,Y)),
    \end{equation*}
    where $d(X,Y)$ denotes the length of the shortest collider free path between $X$ and $Y$. This accounts for the fact that we want to place less weight on incoherencies of tuples that are connected by a long path on which small direct effects could lead to a very low total effect. It can be combined with the Independence-Connection-Mismatch or the Dependence-Separation-Mismatch analogously to the previously presented combined scores.
\end{itemize}
\section{Mini-Review}

\subsection{Recent critisism of causal discovery} \label{app:recent_critisism}
We discuss papers questioning the performance of causal discovery on real data as well as works on the accuracy of score and constraint based causal discovery algorithms.

\paragraph{SCM realism} \cite{jorgensen2025causal} developed a framework to investigate whether the assumption that measured variables follow a causal model can be falsified. They point out that this issue may particularly arise for representations, i.e., a function (e.g., aggregation) of the underlying variables, as illustrated by the classical total cholesterol example, see \cite{jorgensen2025causal}, Example 6.4. This can be seen as a criticism of the assumption that measured variables between which we want to explore relationships follow the logic of SCMs which is a core assumption of many causal discovery methods. The work criticizes what could be described as SCM realism, i.e. assuming observed variables can be adequately modeled by an SCM.

\paragraph{Weak performance on real-world data} There have been improvements in setting up simulational studies to evaluate causal discovery methods with a focus on benchmarking against random guessing, e.g. by \cite{petersen2024you}. However, the question of how well the algorithms perform on semi-synthetic and real-world data remains widely open. \cite{nastl2024causal} tested PC, FCI, and Invariant Causal Prediction (ICP, \citep{peters2016causal}) on real-data and showed that on their selection of 16 real-world data sets, the causal parents found by these methods do not improve out-of-domain prediction compared to using all features as predictors. In favor of the methods, one can point out that \cite{nastl2024causal} tested them on 16 particular datasets for each of which it must be evaluated how well the data conform to the methods' assumptions and if the methods are configured accordingly. Nevertheless, the work indicates that there is a large set of applications for which the methods cannot be applied straightforwardly in their default implementations. Or, as they put it: "Across 16 datasets, we were unable to find a single example where causal predictors generalize better
to new domains than a standard machine learning model trained on all available features." 

\paragraph{Weak performance on semi-synthetic data} Further, \cite{gamella2025causal} introduced the Causal Chambers which generate sensor data from a light and a wind tunnel. In the Chambers, variables can be intervened upon, and the ground truth is known as the underlying physics are well understood. In their case studies, they found that GES \citep{chickering_optimal_2003}, a popular score-based causal discovery method, "struggle[s] with the nonlinear effects of the polarizer angles [...] and the weak effects of the additional light-emitting diodes". For time series data from the wind tunnel, they found that "PCMCI+ \citep{runge_inferring_2019} exhibits low recall and performs similar to random guessing". For PCMCI+, \cite{gamella2025causal} drop edges where conflicts arose. On the one hand, this changes the results drastically and might partially explain the weak performance On the other, this showcases that many PC variants have not implemented appropriate tools for handling orientation conflicts. The case studies of \cite{gamella2025causal} only investigate particular configurations of specific algorithms and are not an exhaustive evaluation of causal discovery. Nevertheless, their work illustrates weaknesses in current causal discovery evaluation and hopefully leads to more extensive benchmarking on semi-synthetic data in the future. 

\subsection{Constraint-based vs. score based causal discovery} \label{app:score_vs_constraint}

Recent work, e.g. \cite{wienobst2025embracing} suggests that score-based methods outperform PC-based methods in discovering DAGs from synthetic data, particularly for linear Gaussian models. The devil's advocate might ask: Why even care to introduce a new evaluation method tailored to error reporting for PC-based causal discovery algorithms? The reason why we aim for better error reporting for PC-based causal discovery methods is twofold. 

\paragraph{Potential for relaxation of assumptions} Synthetic DAG data in the linear Gaussian setting is not the end of the road. Both sets of methods still have to be evaluated more systematically on synthetic, semi-synthetic and real-world data for different set-ups and with varying relaxed assumptions. Better evaluation methods enable the way forward. PC-based methods promise conceptually easier inclusion of hidden common causes, nonlinear relationships and cycles than score-based methods and methods for all these settings are available and used. 

\paragraph{Easy error reporting to and by practitioners needed} Profanely, PC-based methods are, due to their tailoring to different set-ups like stationary time series data or the inclusion of hidden common causes and cycles, widely used by practitioners \citep{brouillard2024landscape} which is in itself a reason for exhaustive error reporting. The presence of a detectable error introduces uncertainty in the entire result that should be reported to and by practitioners.

\subsection{Testing Output Graphs} \label{app:test_graphs}

\cite{ramsey2024choosing} suggest using their proposed Markov checker for hyperparameter tuning of causal discovery methods. Our analysis of error propagation complement their work as our described detectability limits also apply to their setting. Their work also has a conceptual similarity to ours as they test whether the Markov property is fulfilled for a range of graph types and separation criteria while their theoretical focus and use cases differ from ours.

\cite{eulig_toward_2023} mention that their method can be used to falsify output graphs of causal discovery methods with the strong limitation that only CI tests that were neither used to construct the output graph of the algorithm, nor implied via semi-graphoid axioms can be the input for their method.

\section{Experimental results} \label{app:additional_experiments}

For the experiments, we use a parallelized version of PC implemented in the Python package \textit{causy} as the method $\mathcal{M}$. The conditional independence test of choice is Fisher's Z and the threshold $\alpha$ is set to 0.05. Computing the score theoretically takes $\mathcal{O}(n+m)\cdot |\mathcal{T}|$ where $n$ is the number of nodes, $m$ is the number of edges and $|\mathcal{T}|$ is the number of CI tests performed by the PC-like method. Note that we only check separation statements for all tuples tested and do not do further statistical tests. We provide computational times on a MacBook Pro with an Apple M2 Pro Chip and 32 GB memory.

\paragraph{Detecting Incoherencies in the Auto MPG data set.} 
Computing all scores on the auto-mpg data set with six features (variables) and 392 instances (samples) on the computer specified in the introduction of this section took $0.07$ seconds.

The variables $mpg$ and $displacement$ show an ambiguity, in particular, they are tested independent given the three sets $\left[\{acceleration,\, weight\}, \{horsepower, weight\},\{cylinders, weight\} \right]$. Depending on which conditioning set is tested first, $horsepower$ is or is not in the conditioning set for which $mpg$ and $displacement$ are tested independent. As $mpg\,\,\,-\,\,\, horsepower\,\,\,-\,\,\, displacement$ is an unshielded triple in the skeleton, it changes the output graph which one is tested first. This is an ambiguity which is labeled by methods like conservative-PC or resolved by rules like the majority rule. In our case, it can be recovered due to using the parallelized version which does more CI tests than necessary in the classical PC algorithm together with our scores. 

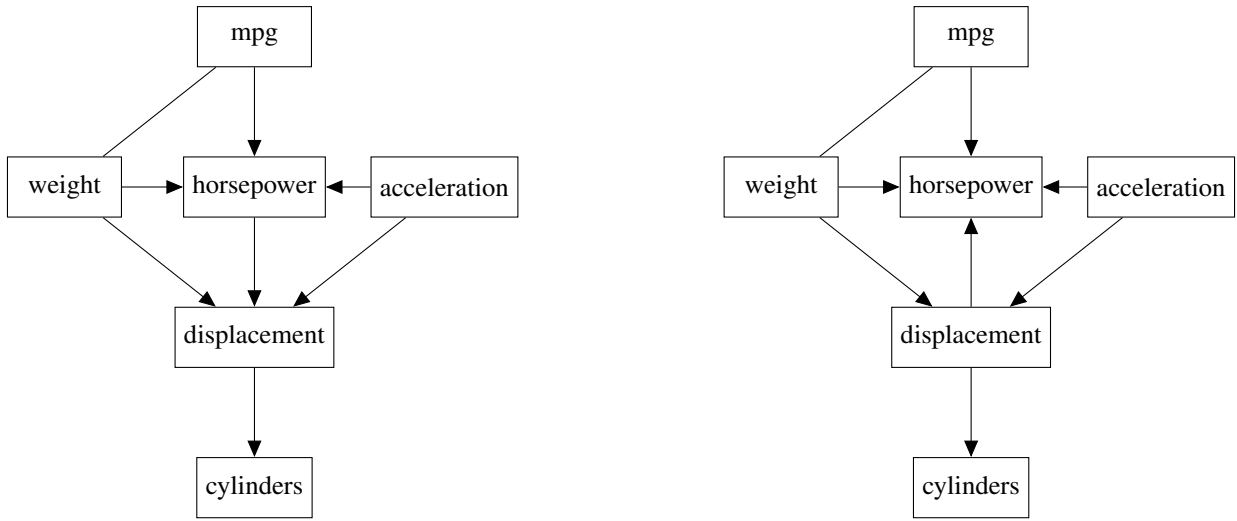
\begin{figure}[h]
    \centering
    \begin{subfigure}{0.45\textwidth}
        \centering
        \begin{tikzpicture}[node distance=2cm, every node/.style={draw, rectangle, minimum width=1.5cm, minimum height=0.8cm}, baseline=(mpg.base)]
            \node (mpg) at (2.5,6) {mpg};
            \node (weight) at (0,4) {weight};
            \node (acceleration) at (5,4) {acceleration};
            \node (horsepower) at (2.5,4) {horsepower};
            \node (displacement) at (2.5,2) {displacement};
            \node (cylinders) at (2.5,0) {cylinders};

            \draw[-] (mpg) -- (weight);
            \draw[->] (mpg) -- (horsepower);
            \draw[->] (horsepower) -- (displacement);
            \draw[->] (displacement) -- (cylinders);
            \draw[->] (acceleration) -- (horsepower);
            \draw[->] (acceleration) -- (displacement);
            \draw[->] (weight) -- (horsepower);
            \draw[->] (weight) -- (displacement);
        \end{tikzpicture}
        \caption{Graph if we lay more emphasis on the test that says that horsepower \textit{is} in the conditioning set yielding that mpg and displacement are independent and \textit{do not} orient the unshielded triple as a collider.}
    \end{subfigure}
    \hfill
    \begin{subfigure}{0.45\textwidth}
        \centering
        \begin{tikzpicture}[node distance=2cm, every node/.style={draw, rectangle, minimum width=1.5cm, minimum height=0.8cm}, baseline=(mpg.base)]
            \node (mpg) at (2.5,6) {mpg};
            \node (weight) at (0,4) {weight};
            \node (acceleration) at (5,4) {acceleration};
            \node (horsepower) at (2.5,4) {horsepower};
            \node (displacement) at (2.5,2) {displacement};
            \node (cylinders) at (2.5,0) {cylinders};

            \draw[-] (mpg) -- (weight);
            \draw[->] (mpg) -- (horsepower);
            \draw[->] (displacement) -- (horsepower); 
            \draw[->] (displacement) -- (cylinders);
            \draw[->] (acceleration) -- (horsepower);
            \draw[->] (acceleration) -- (displacement);
            \draw[->] (weight) -- (horsepower);
            \draw[->] (weight) -- (displacement);
        \end{tikzpicture}
        \caption{Graph if we lay more emphasis on the test that says that horsepower is \textit{not} in the conditioning set yielding that mpg and displacement are independent and \textit{do} orient the unshielded triple as a collider.}
    \end{subfigure}

    \caption{Comparison of output graphs of PC on auto mpg data after ambiguity resolution.}
    \label{fig:auto_mpg}
\end{figure}

We resolved the ambiguity in both ways possible, the output graphs are shown in Figure \ref{fig:auto_mpg}. Table \ref{tab:auto_mpg_coherency_long_version_for_appendix} shows the score results.

\begin{table}[h]
    \centering
    \begin{tabular}{l c c}
        \toprule
         & \textbf{Resolve as collider} & \textbf{Resolve as non-collider} \\
        \midrule
        Total Coherency Score & 0.9174 & 0.9174 \\
        Dependence-Separation-Mismatch (DSM) & 0.9504 & 0.9669 \\
        Independence-Connection-Mismatch (ICM) & 0.9669 & 0.9504 \\
        Conditioning-Set-Size-Adjusted & 0.8895 & 0.9004 \\
        Conditioning Set Size Adjusted DSM & 0.9185 & 0.9358 \\
        Conditioning Set Size Adjusted ICM & 0.9709 & 0.9646 \\
        Number of Orientation Conflicts & 0 & 0 \\
        \bottomrule
    \end{tabular}
    \caption{Coherency scores for PC on auto mpg data with the two possible ambiguity resolutions.}
    \label{tab:auto_mpg_coherency_long_version_for_appendix}
\end{table}

\paragraph{Detecting Incoherencies in the Sachs data set.} The data set contains 11 phosphoproteins and phospholipids in individual cells in each perturbation data set. \cite{sachs_causal_2005} learn a Bayesian network and compare it to expert knowledge. For our configuration, the parallelized version of PC with ParCorr as the CI test returned a graph with orientation conflicts, implying a CI-deviance or a graph-type mismatch. Resolving them naively (strategy: keep first) leads to the a relatively high coherency score of 0.98. Further analysis would be necessary to interpret this result.

\paragraph{Simulational studies: score by SHD to ground truth.} We generated random DAGs with 4 to 9 nodes and $\text{round}(\#\text{nodes}/2)$ to $(\#\text{nodes}\cdot (\#\text{nodes}-1))/2$ edges. For each number of nodes and edges, we drew 100 random DAGs and generated 1000 samples each. We randomly drew edge weights between $0.8$ and $1.2$. We then applied parallel PC and computed the score for each instance. For the method and the computer specified in the introduction of this section, it took 86469 seconds (i.e. 24 hours or 1442 minutes) to compute, where the majority of the computational time is allocated to running a parallel PC for 104000 graphs and only 497 seconds (i.e. 8 minutes) for computing all scores. Computing the total coherency score for one graph took between $8e-05$ seconds (4 nodes, 2 edges) and $0.5$ seconds (9 nodes, 36 edges).

We plotted the mean total coherency scores over 100 instances for each setting in Figure \ref{fig:proxy} by the structural hamming distance (SHD) to the ground truth. We note that we compute the SHD between two CPDAGs: The output of (parallel) PC and the CPDAG of the ground truth DAG from which we sampled. Also, we used the version of SDH for CPDAGs which counts edge differences once, not twice if the edge type differs as it is, for example, also implemented in the Python package \textit{causal-learn}. Our results suggest that the total coherency score on average serves as a proxy for the SHD to the ground truth when the ground truth is not available. Similar to \cite{faller_self-compatibility_2023}, we point out that this can only serve as a heuristic and we do not provide theoretical guarantees –– indeed, we proved one would need further information to derive them. Also, as discussed in the main paper, there can be undetectable CI-deviances leading to a score of one even though the SHD to the ground truth is nonzero as well as detectable CI-deviances leading to a correct output graph, i.e. an SHD of zero.

We also note that the parallel implementation of PC does more tests than necessary due to the parallelizes execution of CI tests with different conditioning sets and conditioning set sizes. The parallelizes version therefore finds more CI-deviances than a classical implementation of PC would.

\paragraph{Assumption violations} For the assumption violation plot in Figure \ref{fig:assumption_analysis}, we drew from the following data-generating SCMs and applied the PC algorithm in the previously describes configuration. For each sample size, we used 100 draws of samples from the SCM and computed the mean score depicted interpolated by a piecewise linear function in the figure, as well as its standard deviation depicted as a transparent environment. All noise terms are standard Gaussian and independent.
\begin{enumerate}
    \item Unconfounded Model, no hidden variables:
    \begin{equation*}
        X=\eta_X,\,\,\, Y=\eta_Y,\,\,\, Z=X+Y+\eta_Z,\,\,\, V=Z+\eta_V,\,\,\, W=Z+\eta_W.
    \end{equation*}
    \item Moderate confounding with hidden variables $\{U_1,U_2\}$:
    \begin{align*}
        U_1&=\eta_{U_1},\,\,\, &U_2&=\eta_{U_2},\,\,\,&X&=0.5\cdot U_1+\eta_X,\,\,\, &Y&=0.5\cdot U_1+\eta_Y,\\  Z&=X+Y+\eta_Z,\,\,\,&V&=Z+0.5\cdot U_2+\eta_V,\,\,\, &W&=Z+0.5\cdot U_2+\eta_W.
    \end{align*}
    \item Strong confounding with hidden variables $\{U_1,U_2\}$:
    \begin{align*}
        U_1&=\eta_{U_1},\,\,\, &U_2&=\eta_{U_2}, &X&=U_1+\eta_X, \quad \quad \quad Y=U_2+\eta_Y,\\ Z&=X+Y+\eta_Z, &V&=Z+U_2+\eta_V,\,\,\, &W&=Z+U_1+\eta_W.
    \end{align*}
    \item Dense confounding with hidden variables $\{U_1,U_2, U_3, U_4\}$:
    \begin{align*}
        U_1&=\eta_{U_1},\,\,\, &U_2&=\eta_{U_2},\,\,\,&U_3&=\eta_{U_3},\,\,\, &U_4&=\eta_{U_4},\,\,\,
        \\
        X&=U_1+U_3+\eta_X,\,\,\, &Y&=U_2+U_4+\eta_Y,\,\,\, &Z&=X+Y+\eta_Z,\,\,\, &V&=Z+U_2+U_3+\eta_V,\\ W&=Z+U_1+U_4+\eta_W.
    \end{align*}
\end{enumerate}
Looking at Figure \ref{fig:assumption_analysis}, we can see that the coherency score is high for the dense confounding model. This is a natural edge case that can be understood by looking at the extreme case: For a model in which all pairs of nodes are confounded, the output graph of the PC algorithm will always be a fully connected graph which matches with the conditional independence results, i.e. $\mathcal{T}^{ind}=\{\}$. Therefore, the coherency score can detect less CI-deviances for densely confounded models. 

The same observation of constant low scores across sample sizes can be made for other detectable assumption violations such as the faithfulness violations induced by the data-generating SCM in Example \ref{ex:unfaithfulness_detectable}, see Table \ref{tab:faithfulness_violation_extended_version_appendix} for mean total coherency scores and the standard deviation across sample sizes (100 repetitions per sample size).
\begin{table}[h]
\centering
\caption{Faithfulness Violation from Example \ref{ex:unfaithfulness_detectable}}\label{tab:faithfulness_violation_extended_version_appendix}\begin{tabular}{ccccc}
\toprule 
\bfseries & \bfseries 50 & \bfseries 100 & \bfseries 1000 & \bfseries 10000 \\ 
\midrule 
Total & 0.9 (0.056)  & 0.879 (0.038)  & 0.876 (0.03)  & 0.884 (0.041)  \\ 
\bottomrule 
\end{tabular}
\end{table}

\paragraph{Method Selection}
Consider the following SCM:
    \begin{align*}
    U_1&=\eta_{U_1}, &U_2&=\eta_{U_2}, &U_3&=\eta_{U_3}, &U_4&=\eta_{U_4} \\
        X&=U_1+U_2+\eta_{X}, &Y&=U_2+U_3+\eta_{Y}, &Z&=U_3+U_4+\eta_{Z}, &V&=U_4+U_1+\eta_{V}.
    \end{align*}
    where $\eta_{X}, \eta_{y}, \eta_{Z}, \eta_{V}, \eta_{U_1}, \eta_{U_2}, \eta_{U_3}, \eta_{U_4} \sim \mathbb{N}(0,1)$ are independent exogenous noise variables and $U_1,U_2,U_3,U_4$ unobserved, see Figure \ref{fig:detectable_causal_insufficiency}. We run the algorithm on the observed variables and note that there is a structural assumption violation, in particular, a causal insufficiency. This is the example discussed in the main paper. On simulated data, we find that the score is almost constant between sample sizes and detects the incoherent results, see Table \ref{tab:causal_insufficiency} for mean scores and standard deviations. It illustrates that the method selection criterion illustrated with oracle-PC on oracle CI statements provided in the main paper translates to versions of PC on simulated data in a natural way. For an undetectable causal insufficiency, see Example \ref{ex:causal_insufficiency_undetectable}.

\begin{table}[h!]
\centering
\caption{Causal Insufficiency}\label{tab:causal_insufficiency}\begin{tabular}{ccccc}
\toprule 
\bfseries Sample Size & \bfseries 50 & \bfseries 100 & \bfseries 1000 & \bfseries 10000 \\ 
\midrule 
Total & 0.891 (0.084)  & 0.846 (0.044)  & 0.843 (0.033)  & 0.849 (0.039)  \\ 
Independence-Connection-Mismatch & 0.896 (0.08)  & 0.846 (0.044)  & 0.843 (0.033)  & 0.849 (0.039)  \\ 
Dependence-Separation-Mismatch & 0.995 (0.037)  & 1.0 (0.0)  & 1.0 (0.0)  & 1.0 (0.0)  \\ 
Orientation Conflicts & 1.27 & 3.14 & 3.58 & 3.44 \\ 
\bottomrule 
\end{tabular}
\end{table}

\begin{figure}[h!]
    \centering
    \begin{subfigure}{0.45\textwidth}
        \centering
        \begin{tikzpicture}[node distance=2cm, every node/.style={draw, circle}, baseline=(y.base), scale=0.6, transform shape]
            \node (x) at (0,0) {X};
            \node (y) at (0,2) {Y};
            \node (z) at (2,2) {Z};
            \node (w) at (2,0) {V};

            \draw[<->] (x) -- (y);
            \draw[<->] (y) -- (z);
            \draw[<->] (z) -- (w);
            \draw[<->] (w) -- (x);
        \end{tikzpicture}
        \caption{The causal graph $\cG_{\text{true}}$ of the data generating SCM. Also, the output of oracle-FCI.}
    \end{subfigure}
    \hfill
    \begin{subfigure}{0.45\textwidth}
        \centering
        \begin{tikzpicture}[node distance=2cm, every node/.style={draw, circle}, baseline=(y.base),scale=0.6, transform shape]
            \node (x) at (0,0) {X};
            \node (y) at (0,2) {Y};
            \node (z) at (2,2) {Z};
            \node (w) at (2,0) {V};

            \draw[dotted, line width=0.5mm] (x) -- (y);
            \draw[dotted, line width=0.5mm] (y) -- (z);
            \draw[dotted, line width=0.5mm] (z) -- (w);
            \draw[dotted, line width=0.5mm] (w) -- (x);
        \end{tikzpicture}
        \caption{Orientation conflicts on all edges in $\cG_{\text{out}}$ after applying oracle-PC.}
    \end{subfigure}
    
    \vspace{0.2cm} 
    \begin{subfigure}{0.45\textwidth}
        \centering
        \begin{tikzpicture}[node distance=2cm, every node/.style={draw, circle}, baseline=(y.base), scale=0.6, transform shape]
            \node (x) at (0,0) {X};
            \node (y) at (0,2) {Y};
            \node (z) at (2,2) {Z};
            \node (w) at (2,0) {V};

            \draw[->] (x) -- (y);
            \draw[->] (z) -- (y);
            \draw[->] (z) -- (w);
            \draw[->] (x) -- (w);
        \end{tikzpicture}
        \caption{The graph $\cG_{\text{out}}$ after applying a conflict resolution strategy: faithfulness incoherent tuple $(Y,V,\emptyset)$.}
    \end{subfigure}
    \hfill
    \begin{subfigure}{0.45\textwidth}
        \centering
        \begin{tikzpicture}[node distance=2cm, every node/.style={draw, circle}, baseline=(y.base), scale=0.6, transform shape]
            \node (x) at (0,0) {X};
            \node (y) at (0,2) {Y};
            \node (z) at (2,2) {Z};
            \node (w) at (2,0) {V};

            \draw[->] (y) -- (x);
            \draw[->] (y) -- (z);
            \draw[->] (w) -- (z);
            \draw[->] (w) -- (x);
        \end{tikzpicture}
        \caption{The graph $\cG_{\text{out}}$ after applying another conflict resolution strategy: faithfulness incoherent tuple $(X,Z,\emptyset)$.}
    \end{subfigure}
    
    \caption{Comparison of outputs of PC on causally insufficient data with/without conflict resolution.}
    \label{fig:detectable_causal_insufficiency}
\end{figure}

\paragraph{Details on Example \ref{ex:very_small_effects_undetectable_erroneous}}

    Consider the following SCM for a constant $c \in \mathbb{R}$:
    \begin{align*}
        X=\eta_{X}, \quad Y=cX+\eta_{Y}, \quad Z=cY+\eta_{Z},
    \end{align*}
    where $\eta_{X}, \eta_{y}, \eta_{Z} \sim \mathbb{N}(0,1)$ are independent exogenous noise variables. This is the model for a simple mediated effect from $X$ to $Z$. We now simulate data from this SCM for different effect strengths $c$, run the PC algorithm and compute the scores for the results. We do so for different sample sizes and average over 100 draws for each sample size. The resulting graphs are shown in Figure \ref{fig:mediated_effect}. The average scores and standard deviations are provided in Table \ref{tab:mediated_effect_02} for $c=0.2$ (most simulations yield the graph in Figure \ref{fig:mediated_effect} (d) as the output graph), in Table \ref{tab:mediated_effect_1} for $c=1$ (most simulations yield the graph in Figure \ref{fig:mediated_effect} (b) as the output graph) and Table \ref{tab:mediated_effect_10} for $c=10$ (simulations with smaller samples often yield the graph in Figure \ref{fig:mediated_effect} (c) as the output, with increasing sample size (b) becomes predominant).
    \begin{figure}[h!]
    \centering
    \begin{subfigure}{0.45\textwidth}
        \centering
        \begin{tikzpicture}[node distance=2cm, every node/.style={draw, circle}, baseline=(y.base)]
            \node (x) at (0,0) {X};
            \node (z) at (3,0) {Z};
            \node (y) at (1.5,1.5) {Y};
            
            \draw[->] (x) -- (y);
            \draw[->] (y) -- (z);
        \end{tikzpicture}
        \caption{The causal graph $\cG_{\text{true}}$ of the data generating SCM.}
    \end{subfigure}
    \hfill
    \begin{subfigure}{0.45\textwidth}
        \centering
        \begin{tikzpicture}[node distance=2cm, every node/.style={draw, circle}, baseline=(y.base)]
            \node (x) at (0,0) {X};
            \node (z) at (3,0) {Z};
            \node (y) at (1.5,1.5) {Y};
            
            \draw[-] (x) -- (y);
            \draw[-] (y) -- (z);
        \end{tikzpicture}
        \caption{Correct result: $\cG_{\text{out}}$ is Markov equivalent to $\cG_{\text{true}}$.}
    \end{subfigure}
    
    \vspace{0.5cm} 

    \begin{subfigure}{0.45\textwidth}
        \centering
        \begin{tikzpicture}[node distance=2cm, every node/.style={draw, circle}, baseline=(y.base)]
            \node (x) at (0,0) {X};
            \node (z) at (3,0) {Z};
            \node (y) at (1.5,1.5) {Y};
            
            \draw[-] (y) -- (z);
        \end{tikzpicture}
        \caption{Markov incoherent tuples in $\cG_{\text{out}}$ due to large effects, e.g. $(X,Z,\emptyset)$.}
    \end{subfigure}
    \hfill
    \begin{subfigure}{0.45\textwidth}
        \centering
        \begin{tikzpicture}[node distance=2cm, every node/.style={draw, circle}, baseline=(y.base)]
            \node (x) at (0,0) {X};
            \node (z) at (3,0) {Z};
            \node (y) at (1.5,1.5) {Y};
            
            \draw[->] (x) -- (y);
            \draw[->] (z) -- (y);
        \end{tikzpicture}
        \caption{Undetectable erroneous result due to effects much smaller than one.}
    \end{subfigure}
    
    \caption{Three output graphs after performing PC on data generated by the SCM in (a) with different effect sizes.}
    \label{fig:mediated_effect}
\end{figure}
\begin{table}[h!]
\centering
\caption{Mediated Effect with $c=0.2$}\label{tab:mediated_effect_02}\begin{tabular}{ccccc}
\toprule 
\bfseries & \bfseries 50 & \bfseries 100 & \bfseries 1000 & \bfseries 10000 \\ 
\midrule 
Total & 0.997 (0.033)  & 0.993 (0.047)  & 1.0 (0.0)  & 1.0 (0.0)  \\ 
Independence-Connection-Mismatch & 1.0 (0.0)  & 1.0 (0.0)  & 1.0 (0.0)  & 1.0 (0.0)  \\ 
Dependence-Separation-Mismatch & 0.997 (0.033)  & 0.993 (0.047)  & 1.0 (0.0)  & 1.0 (0.0)  \\ 
Orientation Conflicts & 0.0 & 0.0 & 0.0 & 0.0 \\ 
\bottomrule 
\end{tabular}
\end{table}

\begin{table}[h!]
\centering
\caption{Mediated Effect with $c=1$}\label{tab:mediated_effect_1}\begin{tabular}{ccccc}
\toprule 
\bfseries & \bfseries 50 & \bfseries 100 & \bfseries 1000 & \bfseries 10000 \\ 
\midrule 
Total & 0.99 (0.057)  & 1.0 (0.0)  & 1.0 (0.0)  & 1.0 (0.0)  \\ 
Independence-Connection-Mismatch & 1.0 (0.0)  & 1.0 (0.0)  & 1.0 (0.0)  & 1.0 (0.0)  \\ 
Dependence-Separation-Mismatch & 0.99 (0.057)  & 1.0 (0.0)  & 1.0 (0.0)  & 1.0 (0.0)  \\ 
Orientation Conflicts & 0.0 & 0.0 & 0.0 & 0.0 \\ 
\bottomrule 
\end{tabular}
\end{table}

\begin{table}[h!]
\centering
\caption{Mediated Effect with $c=10$}\label{tab:mediated_effect_10}\begin{tabular}{ccccc}
\toprule 
\bfseries & \bfseries 50 & \bfseries 100 & \bfseries 1000 & \bfseries 10000 \\ 
\midrule 
Total & 0.703 (0.104)  & 0.74 (0.138)  & 0.973 (0.09)  & 1.0 (0.0)  \\ 
Independence-Connection-Mismatch & 1.0 (0.0)  & 1.0 (0.0)  & 1.0 (0.0)  & 1.0 (0.0)  \\ 
Dependence-Separation-Mismatch & 0.703 (0.104)  & 0.74 (0.138)  & 0.973 (0.09)  & 1.0 (0.0)  \\ 
Orientation Conflicts & 0.0 & 0.0 & 0.0 & 0.0 \\ 
\bottomrule 
\end{tabular}
\end{table}
\section{Additional Examples}
\begin{example} \label{ex:orientation_conflicts}
Let $\mathcal{M}$ be the PC algorithm with marked orientation conflicts. We discuss the output graph in Figure \ref{fig:orientation_conflicts}. We assume that, based on its CI test results, the algorithm has found two colliders $W_1 \rightarrow X_3 \leftarrow W_2$ and $W_3 \rightarrow X_6 \leftarrow W_4$, all other unshielded triples have been classified as non-colliders. According to PC's orientation rules applied to the first collider, all edges between $X_3$ and $X_6$ have to be directed to the right. On the other hand, applying the same rules to the second collider directs these edges to the left, resulting in a conflict which is flagged at one or all of the dotted edges (depending on the implementation). The underlying reason of the conflict is that a DAG with the depicted skeleton for which the set of unshielded colliders is exactly $\{W_1 \rightarrow X_3 \leftarrow W_2, W_3 \rightarrow X_6 \leftarrow W_4 \}$, does not exist. A possible source of this error may be that the collider $W_1 \rightarrow X_3 \leftarrow W_2$ was misclassified and is actually a non-collider. The seemingly unaffected orientations $\{X_3 \rightarrow X_2 \rightarrow X_1$\} are derived from this collider and might thus be incorrect as well.
\end{example}

\begin{figure}[h!]
    \centering
    \resizebox{0.4\textwidth}{!}{
        \begin{tikzpicture}[latent/.append style={minimum size=0.85cm}, obs/.append style={minimum size=0.85cm}, det/.append style={minimum size=1.15cm}, wrap/.append style={inner sep=2pt}, on grid]
            \node[latent] (x) {$X_1$};
            \node[latent, right=2cm of x] (x2) {$X_2$};
            \node[latent, right=2cm of x2] (x3) {$X_3$};
            \node[latent, right=2cm of x3] (x4) {$X_4$};
            \node[latent, right=2cm of x4] (x5) {$X_5$};
            \node[latent, right=2cm of x5] (x6) {$X_6$};
            
            \node[latent, above=2cm of x3] (w1) {$W_1$};
            \node[latent, below=2cm of x3] (w2) {$W_2$};
            \node[latent, above=2cm of x6] (w3) {$W_3$};
            \node[latent, below=2cm of x6] (w4) {$W_4$};

            \draw[->] (w1) -- (x3);
            \draw[->] (w2) -- (x3);
            \draw[->] (w3) -- (x6);
            \draw[->] (w4) -- (x6);

            \draw[->] (x3) -- (x2);
            \draw[->] (x2) -- (x);

            \draw[dotted, line width=0.5mm] (x3) -- (x4);
            \draw[dotted, line width=0.5mm] (x4) -- (x5);
            \draw[dotted, line width=0.5mm] (x5) -- (x6);
        \end{tikzpicture}
    }
\caption{Without conflict resolution, the dotted edges are marked as orientation conflicts, while also the seemingly orientable edges are no longer reliable.}
\label{fig:orientation_conflicts}
\end{figure}

\begin{figure}[h!]
    \centering
    \resizebox{.4\textwidth}{!}{ 
        \begin{tabular}{@{}c@{\hspace{1cm}}c@{}} 
            \begin{tikzpicture}[latent/.append style={minimum size=0.85cm}, obs/.append style={minimum size=0.85cm}, det/.append style={minimum size=1.15cm}, wrap/.append style={inner sep=2pt}, on grid]
                \node[latent] (x) at (0,0) {$X$};
                \node[latent] (y) at (2.4,0) {$Y$};
                \node[latent] (z) at (1.2,2) {$Z$};
                
                \draw[->] (x) -- (y) node[midway, above] {$2$};
                \draw[->] (z) -- (y) node[pos=0.5, right=3pt] {$-2$};
                \draw[->] (x) -- (z) node[pos=0.5, left=4pt] {$1$};
            \end{tikzpicture}
            &
            \begin{tikzpicture}[latent/.append style={minimum size=0.85cm}, obs/.append style={minimum size=0.85cm}, det/.append style={minimum size=1.15cm}, wrap/.append style={inner sep=2pt}, on grid]
                \node[latent] (x) at (0,0) {$X$};
                \node[latent] (y) at (2.4,0) {$Y$};
                \node[latent] (z) at (1.2,2) {$Z$};
                
                \draw[->] (x) -- (z);
                \draw[->] (y) -- (z);
            \end{tikzpicture}
            \\
            $ \cG_{true} $ & $ \cG_{out} $
        \end{tabular}
    }
    \caption{Faithfulness violation leading to a coherent but erroneous result.}
    \label{fig:faithfulness_undetectable}
\end{figure}
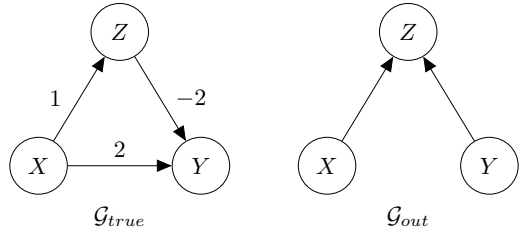

\begin{example} \label{ex:unfaithfulness_undetectable}
    We consider the following ground truth structural causal model:
    \begin{equation*}
        X=\eta_X,\,\,\,\,\,
        Z=X+\eta_Z,\,\,\,\,\,
        Y=2X-2Z+\eta_Y,
    \end{equation*}
    with independent, exogenous noise variables $\eta_X, \eta_Z, \eta_Y \sim \mathcal{N}(0,1)$. The corresponding true causal graph is $\cG_{true}$ in Figure \ref{fig:faithfulness_undetectable}. Note that there is a perfect counteractive mechanism: the direct positive effect of $X$ on $Y$ plus the mediated negative effect of $X$ on $Y$ via $Z$ cancel out, i.e. a faithfulness violation. Assuming no other sources of errors, and choosing the classical PC algorithm as the PC-like method $\cM$, this leads to testing $X$ and $Y$ to be unconditionally independent and all other tuples to be dependent, i.e. $\cT^{ind}=(X,Y,\emptyset)$.
    There is a distribution capturing these independencies and dependencies which is also Markovian and faithful to $\cG^{out}$ (D3), see Figure \ref{fig:faithfulness_undetectable}. There are no orientation conflicts and the method is \textit{not} incoherent (G3). In this case, we have no chance to detect that in fact there was a faithfulness violation and the output graph $\mathcal{G}_{out}$ is erroneous given the list of CI tests and the output graph.
\end{example}

\begin{figure}[h!]
    \centering
        \resizebox{.4\textwidth}{!}{
        \begin{tabular}{@{}c@{\hspace{1cm}}c@{}} 
            \begin{tikzpicture}[latent/.append style={minimum size=0.85cm}, obs/.append style={minimum size=0.85cm}, det/.append style={minimum size=1.15cm}, wrap/.append style={inner sep=2pt}, on grid]
                \node[latent] (x) at (0,0) {$X$};
                \node[latent] (y) at (2.4,0) {$Y$};
                \node[latent] (z) at (1.2,2) {$Z$};
                
                \draw[<->] (x) -- (y);
                \draw[<->] (z) -- (y);
                \draw[<->] (x) -- (z);
            \end{tikzpicture}
            &
            \begin{tikzpicture}[latent/.append style={minimum size=0.85cm}, obs/.append style={minimum size=0.85cm}, det/.append style={minimum size=1.15cm}, wrap/.append style={inner sep=2pt}, on grid]
                \node[latent] (x) at (0,0) {$X$};
                \node[latent] (y) at (2.4,0) {$Y$};
                \node[latent] (z) at (1.2,2) {$Z$};
                
                \draw[-] (x) -- (z);
                \draw[-] (y) -- (z);
                \draw[-] (x) -- (y);
            \end{tikzpicture}
            \\
            $ \cG_{true} $ & $ \cG_{out} $
        \end{tabular}
        }
    \caption{Causal insufficiency leading to a coherent result.}
    \label{fig:causal_insufficiency_undetectable}
\end{figure}
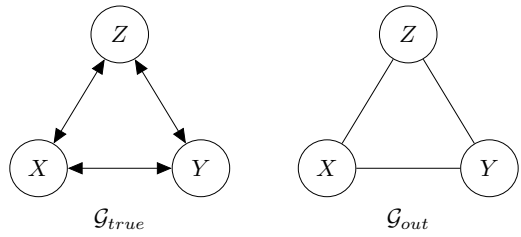

\begin{example} \label{ex:causal_insufficiency_undetectable}
    We want to stress that there can also be undetectable causal insufficiencies. Consider the following SCM:
    \begin{align*}
    U_1&=\eta_{U_1}, &U_2&=\eta_{U_2}, &U_3&=\eta_{U_3},\\
        X&=U_1+U_2+\eta_{X}, &Y&=U_2+U_3+\eta_{Y}, &Z&=U_3+U_1+\eta_{Z}.
    \end{align*}
    where $\eta_{X}, \eta_{y}, \eta_{Z}, \eta_{U_1}, \eta_{U_2}, \eta_{U_3} \sim \mathbb{N}(0,1)$ are independent exogenous noise variables and $U_1,U_2,U_3$ unobserved. The resulting graphs are shown in Figure \ref{fig:causal_insufficiency_undetectable}.
\end{example}

\begin{example} \label{ambiguity_resolution_incoherent}
    As a simple example, take majority-PC as the method $\mathcal{M}$ with one incorrect CI test result leading to an ambiguity and an ambiguity resolution (majority rule) that ignores the one incorrect CI test result. The resulting graph will then be Markov-equivalent to the ground truth causal graph despite one incorrect CI test result.
\end{example}

\begin{example} \label{ex:correct_but_incoherent}
    Consider the classical PC algorithm as the method $\mathcal{M}$ on data generated by the following SCM:
        \begin{equation*}
        X=\eta_X,\,\,\,\,\,
        Z_1=X+\eta_{Z_1},\,\,\,\,\,
        Z_2=Z_1+\eta_{Z_2},\,\,\,\,\,
        Y=Z_2+\eta_Y.
    \end{equation*}
    Now, consider the following CI test results: $\mathcal{T}^{ind}=\{(X,Z_2,\{Z_1\}), (Y,Z_1,\{Z_2\}), (X,Y,\{Z_1,Z_2\})\}$, where the two independencies with conditioning set size one have been missed due to a small sample errors or CI test assumption violations, in particular $(X,Y,\{Z_1\})$ and $(X,Y,\{Z_2\})$. The PC algorithm returns the correct output CPDAG, i.e. the undirected chain $X-Z_1-Z_2-Y$, but the result is CI-deviant. In fact, it is even incoherent, i.e., detectably CI-deviant as $(X,Y,\{Z_1\})$ and $(X,Y,\{Z_2\})$ are tested dependent but d-separated in the output graph.
\end{example}

\section{Further Heuristics and Tests} \label{app:further_tests}
\paragraph{Increasing subsets}
For all assumption violations that lead to incoherencies, the theory implies that for the sample size $n\rightarrow \infty$, the score converges at a score smaller than one. Our simulational studies show that for many settings, this convergence is visible even for moderate sample sizes. These observations motivate the following procedure:
\begin{enumerate}
    \item Compute the score of the method $\mathcal{M}$ on the whole dataset $\mathcal{D}$ as well as for increasing subsets of $\mathcal{D}$.
    \item Analyze whether the scores for increasing subsets stagnate on a low score across sample sizes or if the score increases with increasing sample size.
    \item A stagnating low score might indicate an assumption violation.
\end{enumerate}
We stress that this is a heuristic for error analysis and not a formal statistical test.

\paragraph{Negative control}
A heuristic test to generate a reference value at significance $\beta$. This test can be used to develop a reference value for a particular setting.
\begin{itemize}
    \item[]\textbf{1. Compute score on dataset:} Execute PC-based algorithm $\cM$ with CI test $(T,\alpha)$ on the a dataset $\mathcal{D}$ with sample size $n$. Compute the desired coherency score $sc(\cdot, \cM)$. Record the resulting output graph $\hat{\cG}$ with edge density $p$.
    \item[] \textbf{2. Generate simulated data for comparison:} Generate $K$ (e.g. $K=1000$) Erd\"os-R\'enyi random graphs over the nodes with expected edge density $p$. Generate data from these graphs according to the algorithm’s assumptions. E.g., if the algorithm does not allow for hidden confounding, do not discard variables.
    \item[] \textbf{3. Evaluate:} Run $\cM$ on each simulation and compute score $sc(\cdot, \cM)$ in each run. Reject the null  if the score on the original dataset is worse than $(1-\beta) \cdot 100$ percent of the simulations.
\end{itemize}
\end{document}